%% file: main.tex
\documentclass[10pt,twocolumn,letterpaper]{article}

\usepackage{iccv}
\usepackage{times}
\usepackage{epsfig}
\usepackage{graphicx}
\usepackage{amsmath}
\usepackage{amssymb}
\usepackage{xcolor}
\usepackage{multirow}
\usepackage{booktabs}
\usepackage{subcaption,enumitem}
\usepackage[ruled,vlined,linesnumbered]{algorithm2e}

\newcommand{\PP}[1]{\noindent{\bf #1}}

\usepackage[pagebackref=true,breaklinks=true,letterpaper=true,colorlinks,bookmarks=false]{hyperref}

\usepackage[labelfont=bf,list=true,font=footnotesize, margin=0pt]{subcaption}
\usepackage[labelfont=bf, font=footnotesize, margin=0pt]{caption}
\iccvfinalcopy 

\newcommand{\squishlist}{
        \begin{list}{$\bullet$}
                { \setlength{\itemsep}{0pt}      \setlength{\parsep}{3pt}
                        \setlength{\topsep}{3pt}       \setlength{\partopsep}{0pt}
                        \setlength{\leftmargin}{1.0em} \setlength{\labelwidth}{1em}
                        \setlength{\labelsep}{0.5em} } }

\newcommand{\squishend}{
        \end{list}  }

\def\iccvPaperID{8497} 

\ificcvfinal\fi

\begin{document}

\title{Exploiting Multi-Object Relationships for Detecting Adversarial Attacks in Complex Scenes}

\author{
Mingjun Yin, Shasha Li, Zikui Cai, Chengyu Song, \\
M. Salman Asif, Amit K. Roy-Chowdhury, and Srikanth V. Krishnamurthy \\
University of California, Riverside, USA\\
{\tt\small \{myin013, sli057, zcai032\}@ucr.edu, csong@cs.ucr.edu,} \\
{\tt \small  \{sasif, amitrc\}@ece.ucr.edu, krish@cs.ucr.edu}
}

\maketitle
\ificcvfinal\thispagestyle{empty}\fi

\begin{abstract}
Vision systems that deploy Deep Neural Networks (DNNs) are known to be vulnerable to adversarial examples. Recent research has shown that checking the intrinsic consistencies in the input data is a promising way to detect adversarial attacks (e.g., by checking the object co-occurrence relationships in complex scenes).
{\color{black} However, existing approaches are tied to specific
models and do not offer generalizability.
Motivated by the observation that language descriptions of natural scene images have already captured the object co-occurrence relationships that can be learned by a language model, we develop a novel approach to perform context consistency checks using such language models.
The distinguishing aspect of our approach is that
it is independent of the deployed object detector and yet offers very high accuracy in terms of detecting adversarial examples in practical scenes with multiple objects.
}
Experiments on the PASCAL VOC and MS COCO datasets show that our method can outperform state-of-the-art methods in detecting adversarial attacks.

\end{abstract}

\input{1intro}
\input{2rela}
\input{3method}

\input{4exp}
\input{5conclu}

\noindent \textbf{Acknowledgments.}
{
This material is based upon work supported by the Defense Advanced Research Projects Agency (DARPA) under Agreement No. HR00112090096.
Approved for public release; distribution is unlimited.
}

{\small
\bibliographystyle{ieee_fullname}
\bibliography{egbib}
}
\clearpage
\input{6supp}

\end{document}

%% file: 1intro.tex
\section{Introduction}
\begin{figure*}
    \centering
    \includegraphics[width=0.85\textwidth]{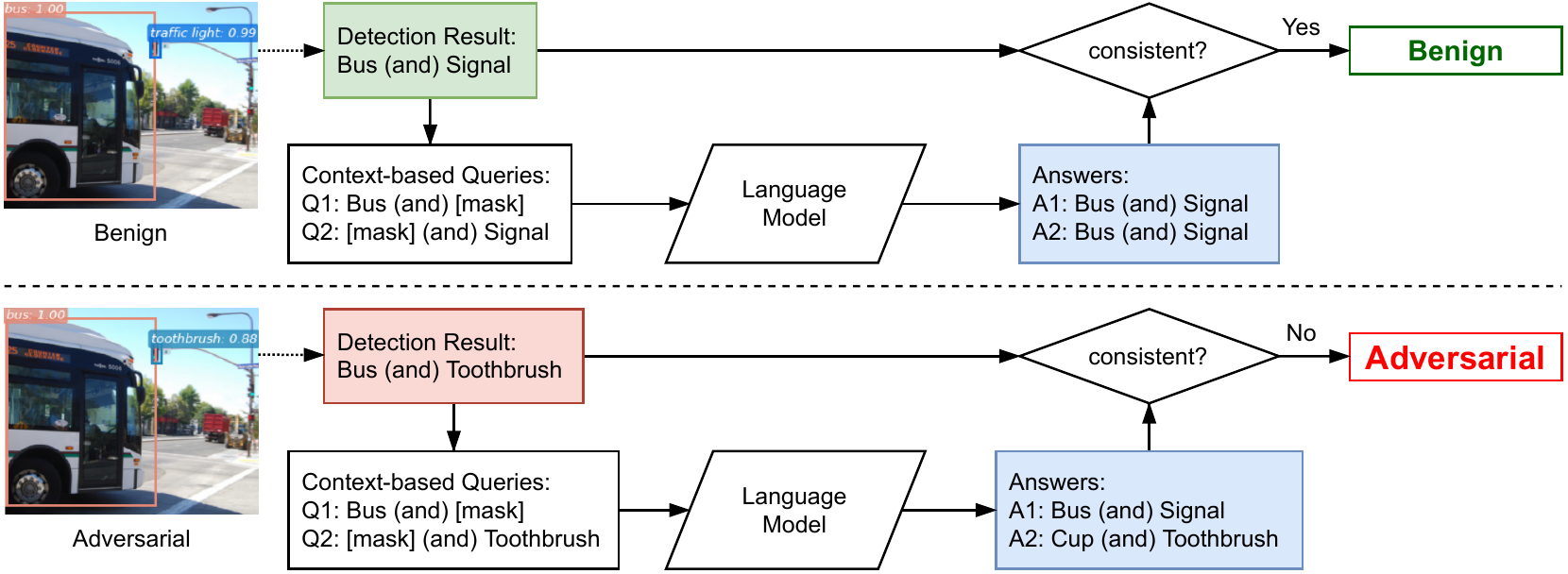}
    \caption{High-level idea of how our language model-based context consistency check works. First, we use a language model to learn the object co-occurrence context (e.g., bus and signal in the example) from descriptions of scene images. At the test time, we mask off detected objects in the scene description and ask the language model to predict the object based on context (i.e., other objects). By measuring the consistency between the detection results and the prediction results, we assess if the input scene image is adversarial or not.}
    \label{fig:motivation}
\end{figure*}

Deep neural networks (DNNs) are widely used in vision tasks such as object detection and classification, for their ability to achieve state-of-the art (SOTA) performance in such tasks.
DNN-based vision systems are also known to be vulnerable to adversarial examples~\cite{goodfellow2018making,szegedy2013intriguing,goodfellow2014explaining,kurakin2016adversarial,carlini2017towards,athalye2018obfuscated,madry2017towards};
specifically, it is possible to add (quasi-)imperceptible perturbations that can cause DNN-based vision systems to output incorrect results, while projecting high confidence with regard to the results.
For example, adversarial examples can  misclassify STOP signs to speed limit signs~\cite{song2018physical} and a school bus to an ostrich~\cite{szegedy2013intriguing}.

One promising defense strategy proposed recently is to capture the intrinsic dependencies within the input data, and to check for violations 
of such dependencies to detect adversarial examples.
For instance, in scene images with multiple objects, the intrinsic relationships between objects, commonly known as the \emph{context} of the scene, can be used to detect adversarial attacks~\cite{li2020connecting}.
Similarly, the dependencies between video frames can be used to detect adversarial frames in video classification~\cite{jia2019identifying,xiao2019advit}.
To illustrate, let us consider the STOP sign attack as an example.
A STOP sign is a part of a road intersection scene wherein it typically co-exists with a stop line and/or a street name sign;
in contrast, a speed limit sign is rarely, if ever, seen at intersections and thus does not co-exist with the latter objects.

While context has been used extensively for object recognition problems and scene understanding, there is little work with respect to detection of adversarial attacks using context. 
{\color{black} In our previous work~\cite{li2020connecting}, we} proposed modeling context as a fully connected graph, where each node is an object proposal from a Region Proposal Network (RPN),
and edges encode how other regions (including the background and the whole scene) affect the current node in its feature space.
Then we train a bank of auto-encoders (each corresponds to a category of objects) to check for consistency with respect to the distribution of context features.
While this approach performs well, it is deeply coupled with Faster R-CNN~\cite{ren2015faster} and cannot be applied to single-stage detectors like YOLO~\cite{redmon2016you}; besides, it requires retraining when there is any change to the Faster R-CNN model (e.g., when switching to another CNN model).
{\color{black}In summary, while prior approaches have tried to utilize context to detect adversarial attacks, they do so in a way that intricately ties the context to the model in use, which  limits their applicability.}

In this paper, we propose a novel {\em model-agnostic} adversarial attack detector based on object co-occurrence.
Our observation is that the language description of a natural scene image (i.e., of the output of an object detection network) can readily capture the dependencies between objects.
We exploit recent advances in natural language models to learn the dependencies between objects based on co-occurrence and to detect adversarial attacks as violations of the learned context model.
\autoref{fig:motivation} depicts the high-level idea of our approach.
Given an unknown scene image, we first encode the output of an object detection network into a sentence describing the object co-occurrence relationships (e.g., ``bus and signals'').
Then we use a trained language model to predict each detected object instance \emph{purely based on the context}.
Finally, we evaluate the context consistency of the scene image by comparing the language model  prediction the and the detection results.
If the results are different, we conclude that the input image is adversarial.

{\color{black}The key contributions of our work are as follows. 
\begin{itemize}[leftmargin=*,noitemsep,topsep=0pt]
\item  To the best of our knowledge we are the first to propose a {\em model-agnostic}, {\color{black} context-consistency based} approach to detect adversarial perturbations against object detectors. 
\item We design and realize a language-based model to learn the object co-occurrence relationships in complex scenes, which serves as our novel context model to detect adversarial attacks. 
\item  We conduct extensive experiments with three different types of adversarial attacks (misclassification, hiding, and appearing) on two large-scale datasets - PASCAL VOC~\cite{everingham2010pascal} and Microsoft COCO~\cite{lin2014microsoft}.
Our method yields high detection performance in all the test cases;
the ROC-AUC is over {\color{black}0.72} in most cases,
which is {\color{black}12-69\%} higher than a state-of-the-art attack detection method~\cite{xu2017feature} that does not use context and
is comparable to (only 5\% worse) previous context-inconsistency-based adversarial attack detection approach~\cite{li2020connecting} that is model-dependent (tightly coupled with the Faster R-CNN architecture thus cannot be applied to other architectures, like YOLO).
\end{itemize}
}

%% file: 2rela.tex
\section{Related Work}
In this section, we review closely related work. 

\noindent \textbf{Object Detection} seeks to locate and classify object instances in images or videos. It is a domain that has been extensively studied~\cite{ren2015faster,liu2016ssd,redmon2016you,lin2017focal}.
Faster R-CNN~\cite{ren2015faster} and YOLO~\cite{redmon2016you} are two state-of-the-art DNN-based object detectors that we considered in this work.
F-RCNN uses a two-stage approach where the first stage proposes bounding boxes and the second stage performs classification.
YOLO takes a single pass design aiming to reducing the computation complexity and improving detection speed.

\noindent \textbf{Context-aware Object Detection} aims to exploit context information to boost the performance of object detection~\cite{oliva2003top,torralba2003contextual,dvornik2018modeling,barnea2019exploring}.
Earlier approaches incorporate context (object co-occurrence) information as a post-processing step to re-score objects detected by DNN-based object detectors~\cite{felzenszwalb2009object,choi2011tree,mottaghi2014role}.
Recent work has also proposed incorporating context as part of the DNN using recurrent units~\cite{liu2018structure} or neural attention models~\cite{hu2018relation}.
{\color{black}Our method takes the post-processing style for (a) the ease of training, and (b) integrating with multiple different object detectors.}

\noindent \textbf{Scene Understanding and Caption Generation} study the problem of generating natural language descriptions for scene images~\cite{xu2017scene,lu2016visual,sadeghi2011recognition,gupta2008beyond,zitnick2013learning,yao2018exploring}.
Besides recognizing objects in a scene, a description generator also needs to detect the relationships or interactions between objects (e.g., ``man riding a horse'').
Although descriptions generated by these systems (e.g., the scene graph~\cite{johnson2018image}) contain richer contextual information and are more discriminative, because predicting correct relationships is a much harder problem than object recognition, existing approaches are not robust enough (i.e., their performance over benign images is not very accurate yet).
For this reason, we opt for the simpler context graph purely based on object co-occurrence where there is only one relationship between objects---they co-occur.
This context can also be described with a simpler language, which is easier to model as well.

\noindent \textbf{Adversarial Attacks against DNNs} are slightly perturbed inputs that can cause a DNN to misbehave~\cite{goodfellow2018making}.
In the visual domain, the perturbations are usually (quasi-) imperceptible noises, but can also be small patches that can be physically applied to a target object~\cite{song2018physical}.
In whitebox settings, adversarial attacks can be generated using gradient-guided optimizations~\cite{szegedy2013intriguing,goodfellow2014explaining,kurakin2016adversarial,carlini2017towards,athalye2018obfuscated,madry2017towards}.
{\color{black} Because our approach takes the prediction results (i.e., labels) as inputs, different attack methods (i.e., how labels are misclassified) will not affect the attack detection results;
so, we only use one attack method in our evaluation.}

\noindent \textbf{Adversarial Attacks against Object Detectors} has received less attention than those against image classifiers.
Most related work focuses on physically realizable attacks~\cite{song2018physical,chen2018shapeshifter,zhao2019seeing}, especially in the domain of autonomous driving vehicles~\cite{cao2019adversarial}.
A key difference from attacking an image classifier is that there are two additional types of attacks viable against object detectors: \textit{hiding} and \textit{appearing} attack~\cite{chen2018shapeshifter,song2018physical}.
Since our attack detection method takes the output of an object detector as input, it is not sensitive to whether the attack is physical or digital.
Therefore, we only evaluate our approach against digital attacks.

\noindent \textbf{Detectors of Adversarial Attacks} aim to distinguish adversarial images from benign ones.
Statistics based detection methods rely on different distributions in the feature space between clean images and perturbed ones~\cite{hendrycks2016early,feinman2017detecting,liu2019detection} to detect adversarial attacks.
Another approach is to transform the input and compare the output of the DNN over the original input with the output over the transformed input; a large inconsistency usually indicates the input is adversarial~\cite{xu2017feature,liang2018detecting}.
For example, Feature Squeeze~\cite{xu2017feature} is a state-of-the-art method that aims to remove useless features from the input space (e.g., by reducing the bit depth of pixels and smoothening surrounding pixels).
We compare with this method in our experiments.

\noindent \textbf{Detecting Adversarial Attacks using Context} is a promising defense strategy explored by recent work.
Xiao et al.~\cite{xiao2018characterizing} propose a detection method {\color{black}in the task of segmentation based on spatial consistency (i.e., how the prediction result of a pixel differs from surrounding pixels).}
Xiao et al.~\cite{xiao2019advit} also proposed using temporal consistency to detect adversarial frames in video clips.
Ma et al., found that the correlation between audio and video can be use to detect adversarial attacks~\cite{ma2019detecting}. 
The closest work to our approach is our earlier work~\cite{li2020connecting}, in which we proposed using an object's context profile, which captures four types of relationships among region proposals (spatial context, object-object context, object-background context, and object-scene context) to detect adversarial perturbations.
Since the context profile used by our previous work is extracted from the internal layers of the object detection network, it is tightly coupled with the object detector.
The approach presented in this work is model-agnostic and thus, does not require expensive retraining to support new object detectors.



%% file: 3method.tex
\section{Methodology}
In this section, we first formalize the problem definition.
Subsequently, we provide an overview of our approach and describe each step in detail.

\begin{figure*}[t]
    \includegraphics[width=\linewidth]{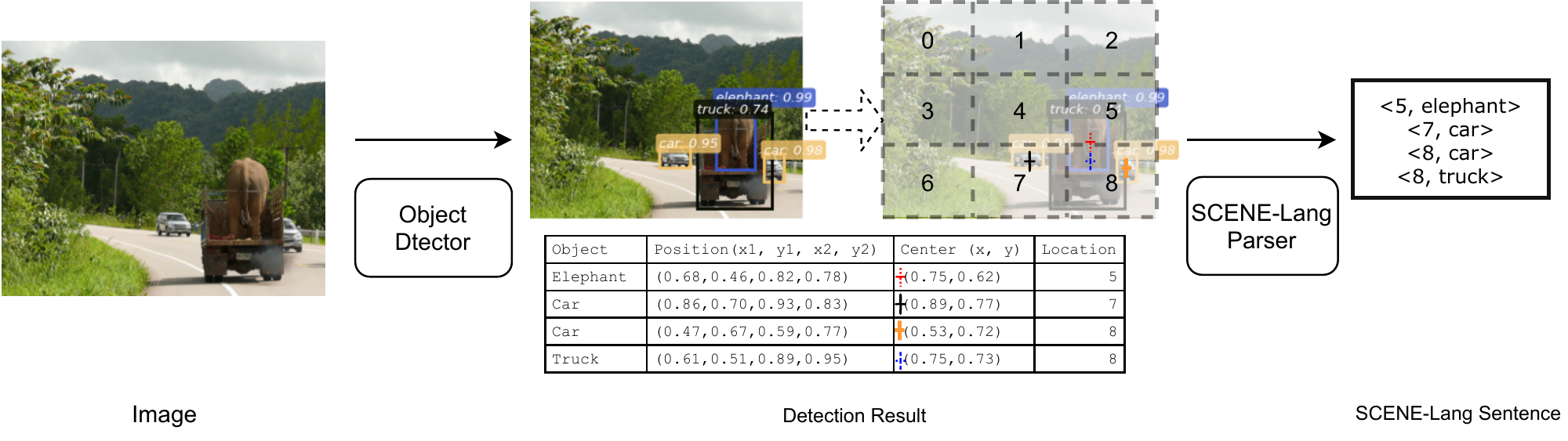}
    \caption{A given image is first processed by an Object Detector (e.g. F-RCNN) to get the detection result. As shown in the table, the positions of object's bounding boxes in the detection results are normalized by the width or height of the image.
For each object, we compute the center of its bounding box. Then, we map the center to an $H \times W$ grid to get a coarse-grained location. Finally, we convert the processed detection results into a SCENE-Lang sentence.
Note that each line in the right-most box is a SCENE-Lang word.
}
    \label{fig:scene-lang}
\end{figure*}

{\bf Problem Definition.}
Let $I$ be a scene image and $C$ be the set of known category labels.
An object detection network $D(I) = O$, takes $I$ as input and outputs a set of detected objects $O=\{(bb_i,c_i),\dots,(bb_n,c_n)\}$,
where $n$ is the number of detected object instances,
$bb_i$ is the bounding box coordinates of the $i$-th instance, and $c_i \in C$ is the category label of the $i$-th instance. 

A co-occurrence context graph $G = (V,E)$ over a set of scene images $\mathcal{I}$ is a fully connected graph,
where a vertex is an object $(bb_i,c_i)$ and the edge between two objects $(bb_i,c_i) \rightarrow (bb_j,c_j)$ encodes the importance of the co-occurrence
({\color{black}i.e., how likely is it that the existence of $(bb_i,c_i)$ 
can predict the existence of $(bb_j,c_j)$}).

The goal of an attacker is to inject a small perturbation to a scene image $I' \leftarrow I + \Delta I$,
so the output of the detection network is manipulated to be $D(I') = O'$ ($O' \neq O$).
Our goal is to determine whether a scene image $I$ is adversarial, with the help of the context graph $G$.

{\bf Threat Model.}
Similar to previous work~\cite{li2020connecting}, we assume a strong white-box attack model where attackers have full knowledge of the object detection network $D(\cdot)$. From an attack detection perspective, this provides defense against the strongest possible attack.
Previous works~\cite{song2018physical,chen2018shapeshifter,zhao2019seeing} have defined three kinds of attacks based on how $O'$ is different from $O$.
\squishlist
 	\item {\em Misclassification attack}, where the label of an instance is misclassified, i.e., $c'_i \neq c_i$;
 	\item {\em Hiding attack}, wherein an instance is not detected by the victim object detector, i.e., $(bb_i,c_i) \notin O'$;
 	\item {\em Appearing attack}, where an instance that does not exist, is detected by the victim object detector, i.e., $(bb'_i,c'_i) \notin O$.
\squishend

{\bf Overview.}
Our approach uses context consistency checks to detect adversarial attacks,
where the context is defined by the co-occurrence of objects within the scene and their relative positions.
The two main challenges in realizing this approach are:
(1) how to learn the context graph $G$ (i.e., the edge weights), and
(2) given a test time co-occurrence relationship $O$, how to check whether it is consistent with $G$.

In this work, we explore the feasibility of using natural language models to solve these challenges.
In particular, we first define a new language {\tt SCENE-Lang} to capture the category and coarse-grained location of object instances in a scene image. 
We can then \emph{describe} the output $O$ of an object detection network using a \emph{sentence} in SCENE-Lang.
Such scene descriptions (i.e. sentences in SCENE-Lang) form the underpinning for training a language model that essentially models the context graph $G$.
In this work, we use a model based on BERT~\cite{devlin2019bert, liu2019roberta}, which we call SCENE-BERT, to learn the intrinsic dependencies between words (i.e., co-occurring objects).
{Compared to alternative methods like 2-D co-occurrence matrix, graph neural networks, and message-passing-based RNNs, we believe that the attention mechanism of the BERT model allows the capture of dependencies between objects with significantly reduced computation. We also believe that BERT will perform better on naturally occurring, more complicated scenes (i.e., images with more objects) because they can appropriately activate the attention heads from transformers.}

During test time, we generate a context consistency score for a scene image $I$ based on the trained SCENE-BERT model.
A violation of context consistency in the scene image will lead to a low consistency score.
This allows us to detect adversarial attacks by thresholding the consistency score of the composed SCENE-Lang sentence from the scene image.
The overall workflow is depicted in~\autoref{fig:motivation}. 
We point out here (again) that because SCENE-BERT learns the context from the output of an object detection network $D(\cdot)$,
it can work with most detection networks like F-RCNN and YOLO.
More importantly, because SCENE-BERT can be trained independently (e.g., using the ground truth labels),
applying it to a new detection network neither requires the retraining of the detection network nor SCENE-BERT itself.

\subsection{Learning Context with Language Model}

{\bf SCENE-Lang.}
We define a new language called SCENE Language (SCENE-Lang) to describe the object co-occurrence information in natural scene images.
Each natural scene image can be described with one SCENE-Lang sentence and each word in the sentence is associated with an object instance. 

{\em SCENE-Lang Words.}
We describe the category of an object and its coarse-grained location with a SCENE-Lang word. 
To describe the location of an object, we evenly divide each image into a $H\times W$ grid and label each grid cell using a number{, so we can use a small, finite vocabulary to describe the scene.
Using coarse-grained locations can also help tolerate adversarial attacks that may shift the bounding boxes of objects. }
The center of an object's bounding box determines which cell the object is in.
We denote the set location labels as $L = \{1,\dots,H\times W\}$.
Therefore, each SCENE-Lang word $w = (l,c)$ is a pair of a location label ($l \in L$) and a category label ($c \in C$).
We denote the finite vocabulary of SCENE-Lang as $W = C \times L$, whose size $|W|=|C| \times H \times W$.
{\color{black}Note that although we could  encode the object label $c_i$ using a number, because SCENE-Lang is a pseudo language
we choose to use the natural language label $c_i$; this enables ease of explanation upon detecting a context consistency violation.}

{\em SCENE-Lang Sentence.}
We describe the object co-occurrence relationships in a scene image $I$ with a single SCENE-Lang sentence, wherein each word is associated with an object instance in the image.
The sentence is represented by $s_I = [w_1,\dots,w_n]$, where the length of the sentence $n$ is equal to the number of object instances in the image $I$.
We use $s$ instead of $s_I$ subsequently, for ease of exposition.
The order of the words in the sentence is sorted based on their location labels (numerically ascending).
\autoref{fig:scene-lang} shows an example about how to describe a scene image with a SCENE-lang sentence.
 
{\bf SCENE-BERT.}
We use SCENE-BERT, a natural language model to learn the co-occurrence context graph $G$ in natural scene images. 
Each input to SCENE-BERT is a sequence of tokens, denoted as $\mathbf{T}=[t_1,\dots,t_n]$.
The model also takes a n-dimensional mask vector $\mathbf{M}\in \{0,1\}^{n}$ as input, where $0$ in the $i$-th dimension indicates masking off the $i$-th token $t_i$
and $1$ indicates that the corresponding token $t_i$ is not masked.
The number of tokens $n$ is determined by the number of words in the input SCENE-Lang sentence.
The output of SCENE-BERT is a reconstructed sequence of tokens,
where the masked off token in $\mathbf{T}$ is replaced with a listed of predicted tokens that match the context (i.e., with lowest cross-entropy loss).
We use $f(\mathbf{T},\mathbf{M},t_i)$ to denote the confidence score of $t_i$ in the predicted token list.
If $t_i$ is not in the list, $f(\mathbf{T},\mathbf{M},t_i) = 0$.
This score will be used to calculate the consistency score of the whole scene.

\begin{figure}
    \includegraphics[width=0.45\textwidth]{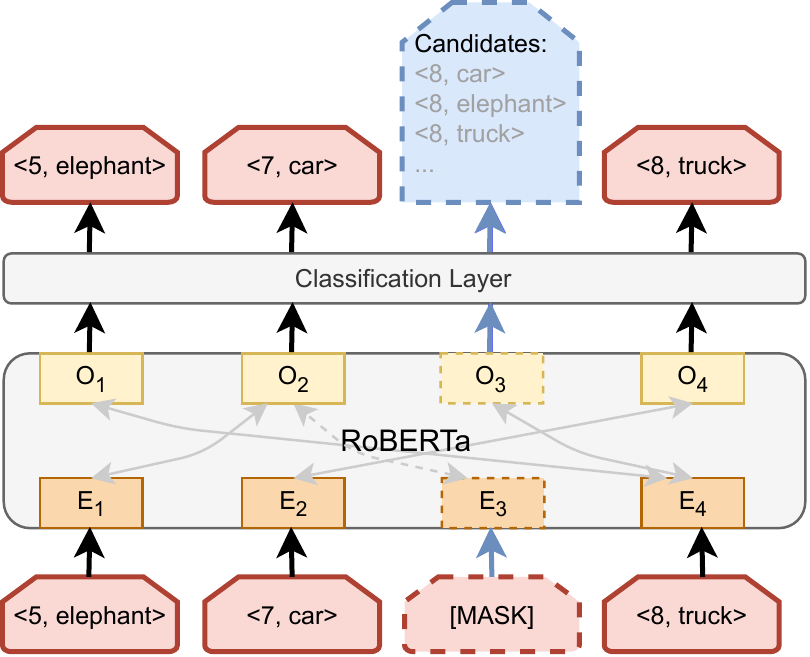}
    \caption{During training or testing time, we selectively mask one or some tokens and ask the SCENE-BERT to predict the masked part. We update weights based on the prediction result at training time. We take the prediction results as the categories, which fit in current context, at test time. } 
    \label{fig:scene-mlm}
\end{figure}

{\em SCENE-BERT Architecture.}
SCENE-BERT is based on the multi-layer bidirectional transformer model BERT~\cite{devlin2019bert}.
\autoref{fig:scene-mlm} shows the simplified architecture of the model.
Since transformers have been widely adopted in language-related tasks and we are reusing an existing implementation of BERT,
we omit the detailed description of the model architecture and refer the readers to~\cite{vaswani2017attention}, in the interest of space.
Besides being a state-of-the-art language model, we choose BERT to implement our language model for two main reasons.
First, the use of the bidirectional self-attention mechanism allows BERT to capture dependencies from both directions
(i.e., the prediction of the current word depends on both words appearing before it and after it).
{\color{black}This matches our context model very well as the context graph $G$ is not ordered (relationships are both ways).}
Second, the way BERT is trained is also a perfect match for our task.
In particular, BERT is trained with the Masked Language Modeling (MLM) task, where some input tokens are masked at random and the model is asked to predict them.
This task is very similar to our approach to detect adversarial attacks (\autoref{fig:motivation}): check whether object detection result is consistent with those predicted purely based on context.

{\em Tokenization.}
Because SCENE-Lang is a pseudo language, tokenizing a sentence $s$ is straightforward.
Specifically, we assign each unique word $w$ in the finite vocabulary of SCENE-Lang $W$, a unique number,
which serves at its corresponding token (i.e., $t \in \{1,\dots,|W|\}$).
So the tokenizer simply maps each word $w_i$ in a sentence $s$ to its corresponding number.


{\em Training.}
We train SCENE-BERT using the same unsupervised masked language modeling task as RoBERTa~\cite{liu2019roberta}.
Specifically, it randomly masks token(s) from the input sequence, and the objective of the model is to predict the original token(s), only based on the remaining tokens in the sentence (i.e., the context).
In other words, SCENE-BERT learns the dependencies between co-occurring objects, or the edge weights in the object co-occurrence graph $G$.
We want to highlight the unique advantage of SCENE-BERT again in that it can be trained with any set of sentences in SCENE-Lang.
{\color{black} This means that it can be trained with the ground truth labels of an object detection dataset (as we do in our experiments); this would work with any object detection network trained with the same dataset.
Alternatively, it can also be trained in a completely unsupervised manner by running an object detector over a clean dataset to generate the training sentences.}

\subsection{Checking Context Consistency}
In this subsection, we illustrate how we use the trained SCENE-BERT model to perform context consistency checks.
At a high level, we use differential analysis to detect inconsistency, i.e., by comparing the detection result (in SCENE-Lang) with the scene description predicted from our context model SCENE-BERT.
The smaller the difference, the higher will be the consistency score.
Because most adversarial attacks will violate context consistency, by thresholding the consistency score, we can detect whether the input image is adversarial or not.
Next, we introduce how we calculate the consistency score.

%

Let $I$ be a clean scene image and $D(\cdot)$ be the victim object detector.
We can encode the output $D(I) = O$ using a SCENE-Lang sentence $s = [w_1,\dots,w_n]$ as described earlier,
which will be tokenized into $\mathbf{T} = [t_1,\dots,t_n]$.
Let $I' \leftarrow I + \Delta I$ be the perturbed adversarial image and $\mathbf{T}'$ be the tokenized SCENE-Lang description over $I'$.
Recall that there are three possible attacking goals, which will affect $x$ in three different ways:
\squishlist
 	\item Misclassification attack where the token associated with an instance is perturbed, i.e., $t'_i \neq t_i$;
 	\item Hiding attack where a token is missing in the token sequence, i.e., $t_i \notin \mathbf{T}'$;
 	\item Appearing attack where an undesired token appears, i.e., $t'_i \notin \mathbf{T}$.
\squishend

Using the trained SCENE-BERT model, we can mask off a token $t'_i \in \mathbf{T}'$ and ask the model to predict what $t'_i$ is, based on the remaining tokens (i.e., the test time context).
In theory, if the predicted result is different from $t'_i$, then we deduce that $t'_i$ is the likely target object under attack.
However, this has two associated problems.
First, there could be multiple objects that are contextually consistent (i.e., SCENE-BERT can return a list of possible tokens instead of a single one),
and hence, how should we calculate the difference between $t'_i$ and the predicted tokens?
Second, $\mathbf{T}'$ contains multiple tokens and so, how can we know which token to mask, especially in the case of a hiding attack (where the victim token is missing)?
We solve the first problem by using the confidence score of $t'_i$ in the predicted list as the consistency score of that specific object.
If $t'_i$ is not in the predicted list from SCENE-BERT, its consistency score will be $0$.
We solve the second problem by iterating through all tokens (i.e., detected objects) and using the lowest consistency score of all objects as the consistency score of the whole image.
The details are captured in Algorithm~\ref{algo:score}.
{\color{black}Note that our approach to calculate the consistency score is able to handle hiding attacks because the missing token typically affects the prediction results of the other non-target tokens.}

\begin{algorithm}[t]
	\fontsize{8}{8}
	\selectfont
	\SetAlgoLined
	\SetKwInOut{Input}{Input}\SetKwInOut{Output}{Output}
	
	\Input{Tokenized SCENE-Lang sentence $\mathbf{T}=[t_1,\dots,t_n]$,
	the trained SCENE-BERT function $f(\cdot,\cdot,\cdot)$}
	\Output{Consistency score $c$}
	$c = 1.0$ \\ 
	$\mathbf{M} = {1}^n$ \\ 
	\For {$i=1 \text{ to } n $}{
		$\mathbf{M}[i] = 0$\\
		$r_{t_i} \gets f(\mathbf{T},\mathbf{M},t_i)$ \\
		$c = \min(c, r_{t_i})$ \\
		$\mathbf{M}[i] = 1$\\
	}
	\Return $c$
	\caption{Calculate the consistency score of a SCENE-Lang sentence.}
	\label{algo:score}
\end{algorithm}

%% file: 4exp.tex
\section{Experimental Analysis}
In this section, we evaluate the performance of our approach through comprehensive experiments on two large-scale object detection datasets:  PASCAL VOC~\cite{everingham2010pascal} and MS COCO~\cite{lin2014microsoft}.
We used the two most popular object detection networks: Faster R-CNN~\cite{ren2015faster} and YOLO~\cite{redmon2016you}.
We also compare our approach with two state-of-the-art adversarial attack detection methods: a context-agnostic one feature squeeze~\cite{xu2017feature} and another context-aware detection method {\tt SCEME}~\cite{li2020connecting}.
The evaluation includes three types of attacks: misclassification, hiding, and appearing.

\subsection{Implementation Details}
\label{sec:impldetails}

We use the RoBERTa~\cite{liu2019roberta} model, a reproduction of the original BERT model~\cite{devlin2019bert}, to implement SCENE-BERT.
It is configured with six hidden layers and twelve self-attention heads.

The PASCAL VOC dataset contains 20 object categories. The majority of images in the PASCAL VOC dataset have 1 to 5 object instances, on average, 1.4 categories and 2.3 instances per image.
The MS COCO dataset contains 80 object categories. Images in this dataset have more object instances, on average, 3.5 categories and 7.7 instances per image.
We used the ground truth labels from both datasets to train SCENE-BERT models.
{Since our context model is designed to consider the co-occurrence consistency of multiple objects in the scene, we omitted images that consist of a single object.}
For the PASCAL VOC 2007 dataset, we used a $3\times 3$ grid (i.e., $H=3$ and $W=3$); so in total we have $|W|=C\times H \times W=20\times 3\times 3=180$ tokens. 
%
For MS COCO, we also used a $3\times 3$ grid, so in total $|W|=C\times H \times W=80\times 3\times 3=720$ tokens.

Since SCENE-BERT can be trained independent of the object detector, we used pre-trained F-RCNN and YOLO models. 
For the PASCAL dataset, both models were trained with \textit{VOC07trainval} and \textit{VOC12trainval}.
For the MS COCO dataset, the F-RCNN model was trained with \textit{coco14train} and \textit{coco14valminusminival}, and the YOLO model was trained with \textit{coco17train}.

To test the attack detection performance, we generate 10,000 attacks for each attacking goal (misclassification, hiding, and appearing) from both datasets, except for hiding attacks on PASCAL VOC, which does not have enough objects for hiding attacks.
Because our detection method uses high level semantic information (object co-occurrence context) and does not rely on low-level features, we only evaluate it against digital attacks.
The attacks are generated using the standard iterative fast gradient sign method (IFGSM)~\cite{kurakin2016adversarial}, with $L_{\infty} \leq 10$ as the perturbation budget; and the perturbations are applied to the whole image.
{\color{black}Because SCENE-BERT takes detected object labels and locations as inputs, how the perturbations are generated {does not affect the experimental analysis}, thus we only used IFGSM.}
\autoref{tbl:attacks} shows the attack success rate on the two datasets.

\begin{table}[t]
    \centering
    \footnotesize
    \caption{Attack success rate of three different goals on the PASCAL VOC and MS COCO datasets.}
    \label{tbl:attacks}
    \begin{tabular}{cccc}
    \toprule
    Model & Misclassification & Hiding & Appearing \\
    \midrule
    \multicolumn{4}{l}{\textbf{Results on PASCAL VOC:}} \\
    \midrule
    F-RCNN & 90.33\% & 78.09\% & 96.01\% \\ 
    YOLO  & 80.16\% & 89.03\% & 94.78\% \\
    \midrule
    \multicolumn{4}{l}{\textbf{Results on MS COCO:}} \\
    \midrule
    F-RCNN & 92.78\% & 82.34\% & 94.49\% \\ 
    YOLO  & 79.82\% & 93.77\% & 89.74\% \\ 
    \bottomrule
    \end{tabular}
\end{table}


\subsection{Baseline Models}
We compare our method with two baseline models in the experiments.

{\bf Feature Squeeze} (FS)~\cite{xu2017feature} is a SOTA context-agnostic method for detecting adversarial image examples.
This mechanism can detect the adversarial image examples generated by Fast Gradient Sign Method~\cite{goodfellow2014explaining}, DeepFool~\cite{moosavi2016deepfool}, and Projected Gradient Descent~\cite{madry2017towards}. 
{Its core idea is that, adversarial attacks need to limit how many perturbations can be applied (e.g., by limiting the change to the $L_2$ or $L_\infty$ norm) to achieve (quasi-)imperceptibility.
Therefore, by squeezing the input features (i.e., reducing the color bit depth of each pixel and smoothing surrounding pixels), FS may remove enough perturbations and acquire the correct prediction results.
Then, by comparing how different the prediction results of the original input and the squeezed input are, FS can detect adversarial attacks.
}
%

{\bf SCEME}~\cite{li2020connecting} is our previous context-consistency-based adversarial attack detection method that showed much better detection performance than Feature Squeeze.
It models context at region proposal level and uses attention mechanism and Gated Recurrent Units (GRUs) to learn four types of relationships between region proposals:
(1) \textit{spatial context} between regions corresponding to the same object;
(2) \textit{object-object context} between regions corresponding to different objects;
(3) \textit{object-background context} between regions corresponding to objects and regions corresponding to the background; and
(4) \textit{object-scene context} between regions and the whole scene.
To detect adversarial attacks, SCEME uses auto-encoders (one per object category) to learn the benign distribution of \emph{context profiles} corresponding to an object category.
The context profile contains both edge features and node features (i.e., features of the region proposal).
An adversarial attack that violates context consistency will yield a higher reconstruction error rate and by thresholding the reconstruction error rate, SCEME can detect perturbed \emph{regions}.
Note that because SCEME works at region proposal level instead of the whole image, we cannot directly compare it with SCENE-BERT. 
To calculate the detection performance at the whole image level, we aggregate all reconstruction errors from each region proposal and use the highest one as the final score.

\begin{figure*}[t]
    \centering
    \begin{subfigure}{0.32\linewidth}
        \centering
        \includegraphics[width=\textwidth]{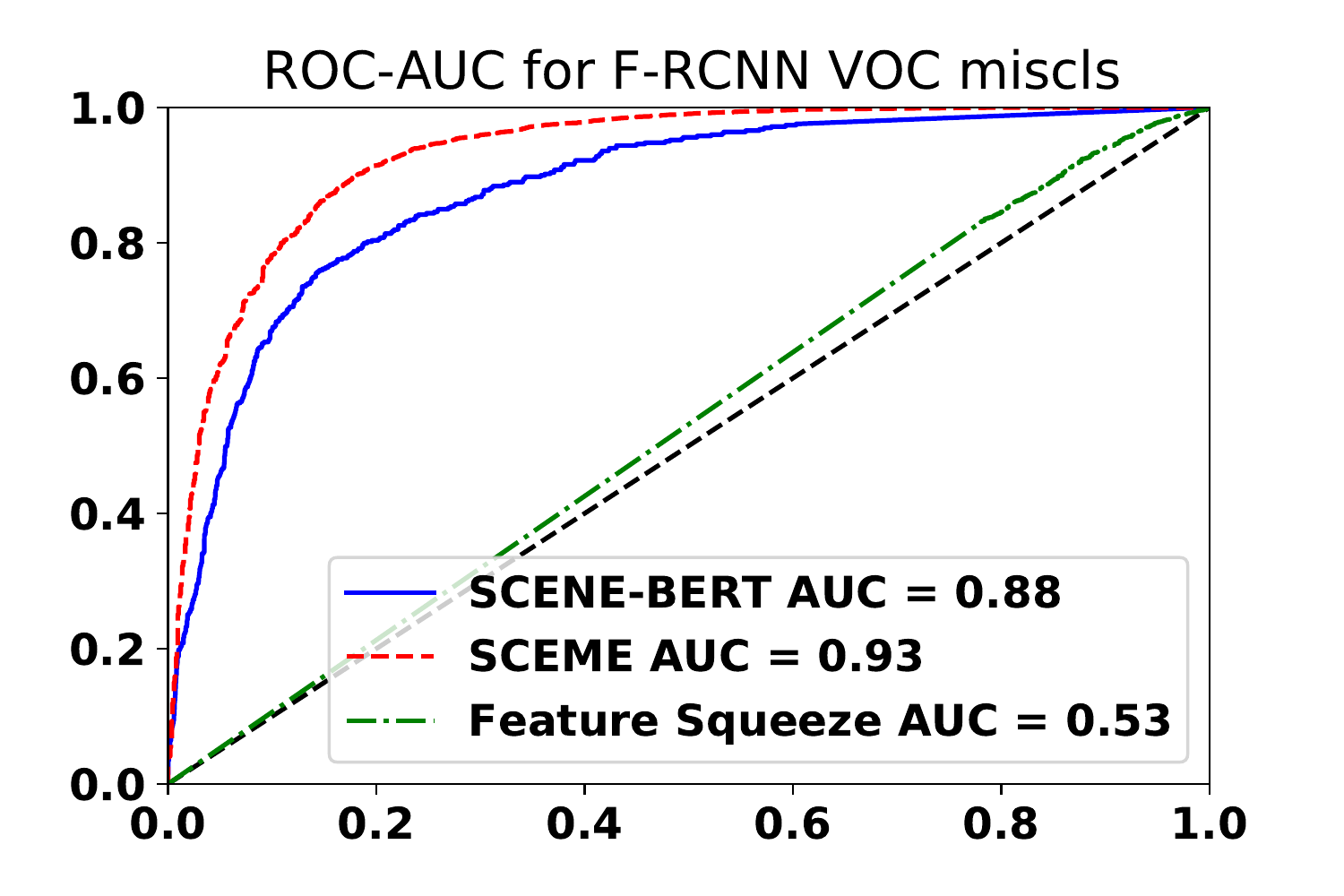}
        \label{fig:F-RCNNvoc}
    \end{subfigure}
    \begin{subfigure}{0.32\linewidth}
        \centering
        \includegraphics[width=\textwidth]{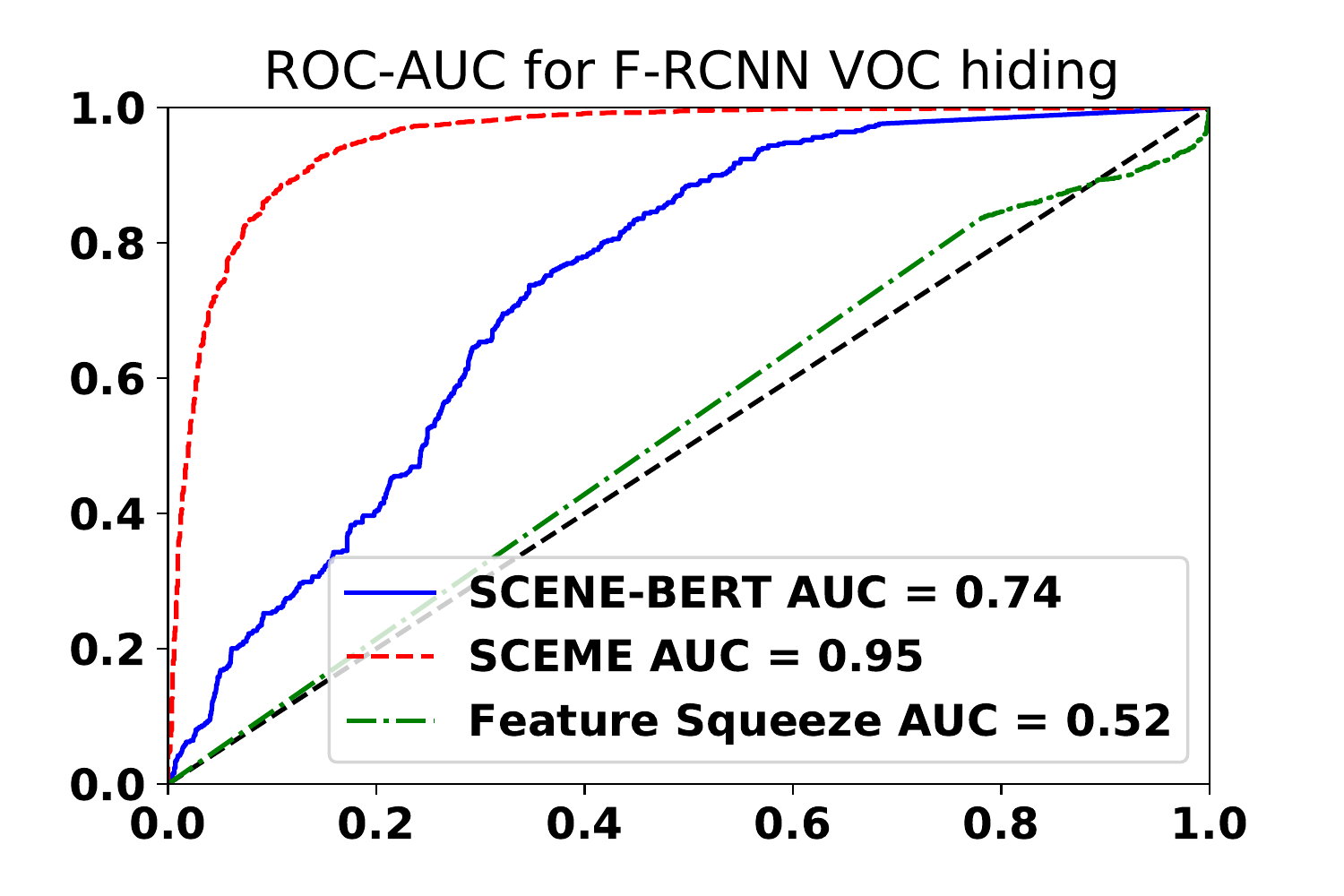}
        \label{fig:F-RCNNvoc}
    \end{subfigure}
    \begin{subfigure}{0.32\linewidth}
        \centering
        \includegraphics[width=\textwidth]{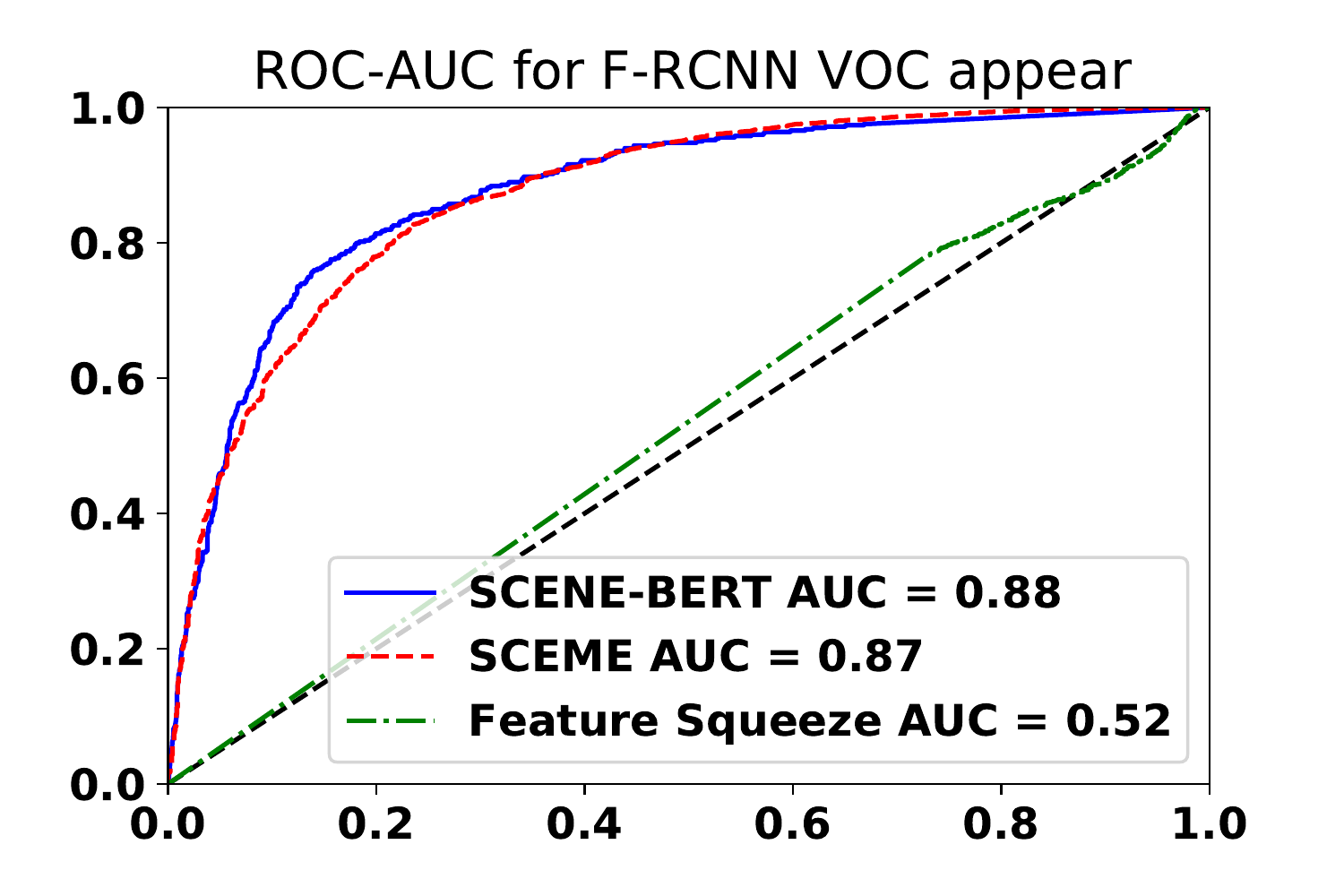}
        \label{fig:F-RCNNvoc}
    \end{subfigure}
    \caption{Detection performance on the PASCAL VOC dataset.}
    \label{fig:voc}
\end{figure*}

\subsection{Detection Performance}\label{sec:results}

\PP{Evaluation Metric.}
Given a scene image and an object detector, we aim to determine whether the scene image is adversarial (i.e., the object detector is fooled by the image and makes a wrong prediction). We first compose the SCENE-Lang sentence using the detection result of the scene image output by the object detector. We then use the SCENE-BERT model to calculate the consistency score of the composed SCENE-Lang sentence. We expect that benign/negative images have higher consistency scores and adversarial/positive images have lower consistency scores. By thresholding the consistency score, we are able to plot the receiver operating characteristic (ROC) curve  of the detection. We report the area under the curve (AUC) of the ROC curve to evaluate the detection performance.


\input{tables/results}

\PP{Detection Performance.}
\autoref{tab:result} shows the detection performance on the PASCAL VOC dataset and the MS COCO dataset.
{\autoref{fig:voc} visualizes the AUC curves on the PASCAL VOC dataset with F-RCNN under different attack setups for better comparison with SCEME and FS.}
Overall, SCEME and SCENE-BERT, both of which are the context-aware detection methods, significantly outperformed Feature Squeeze, which is a context-agnostic method.
The only exception is hiding attacks on the MS COCO dataset.
The reason is that images from the MS COCO dataset have more objects, so hiding a single object usually will not significantly reduce the context consistency.
We believe the results once again validate the effectiveness of context consistency-based detection approach.
Comparing SCEME and SCENE-BERT, we observed that SCEME still outperformed SCENE-BERT.
We attribute this to the richer features used by SCEME (e.g., object-background and object-scene context).
However, SCENE-BERT also has its advantages over SCEME.
First, SCENE-BERT is model-agnostic, so we can also pair it with YOLO without any modification to YOLO or retraining;
on the other hand, SCEME is tightly coupled with the Faster R-CNN architecture.
Second, SCENE-BERT is also faster as it only iterates through detected object instances, whereas SCEME needs to iterate through hundreds of region proposals.

\begin{figure*}[t]
    \centering
    \begin{subfigure}{0.32\linewidth}
        \centering
        \includegraphics[width=\textwidth]{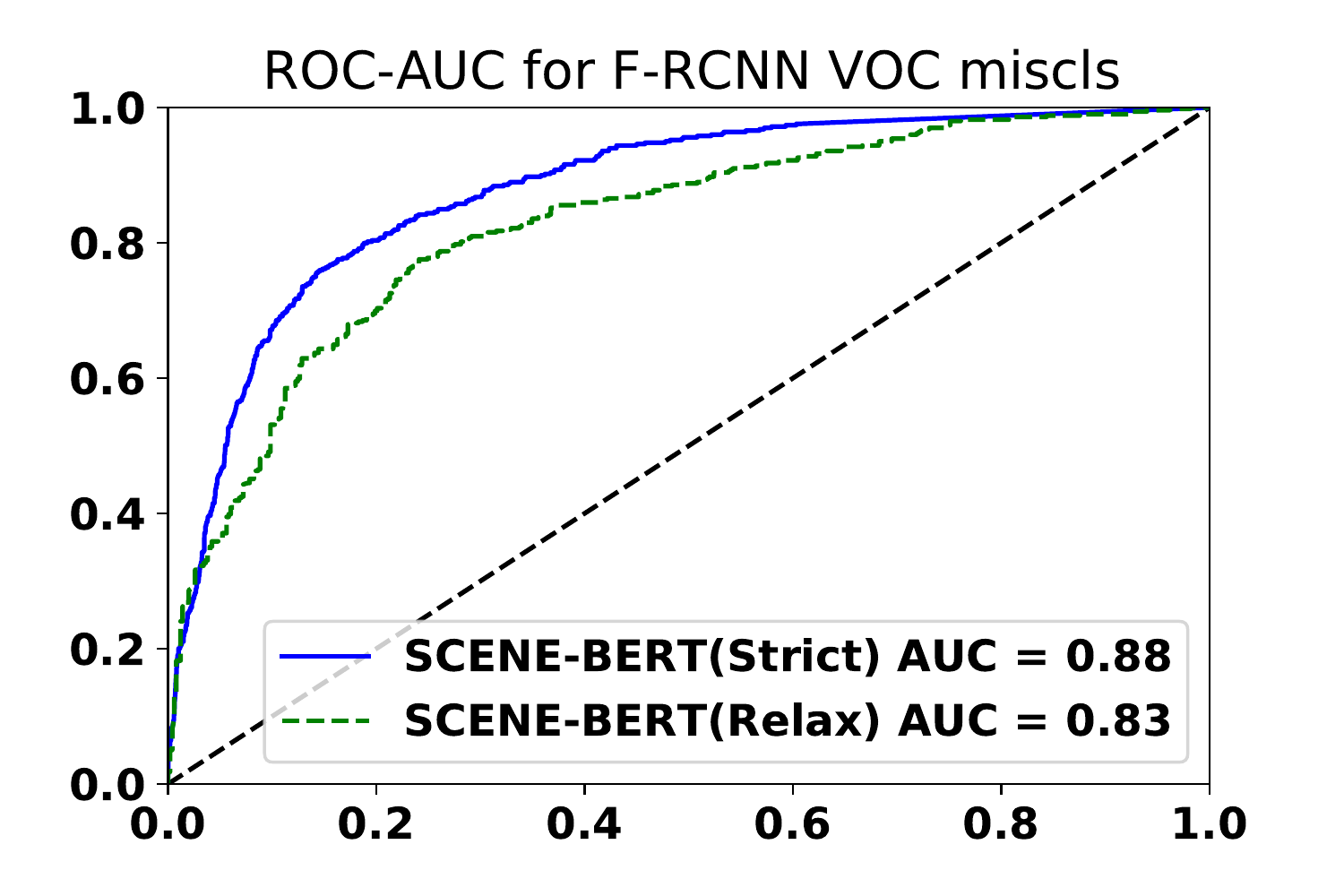}
    \end{subfigure}
    \begin{subfigure}{0.32\linewidth}
        \centering
        \includegraphics[width=\textwidth]{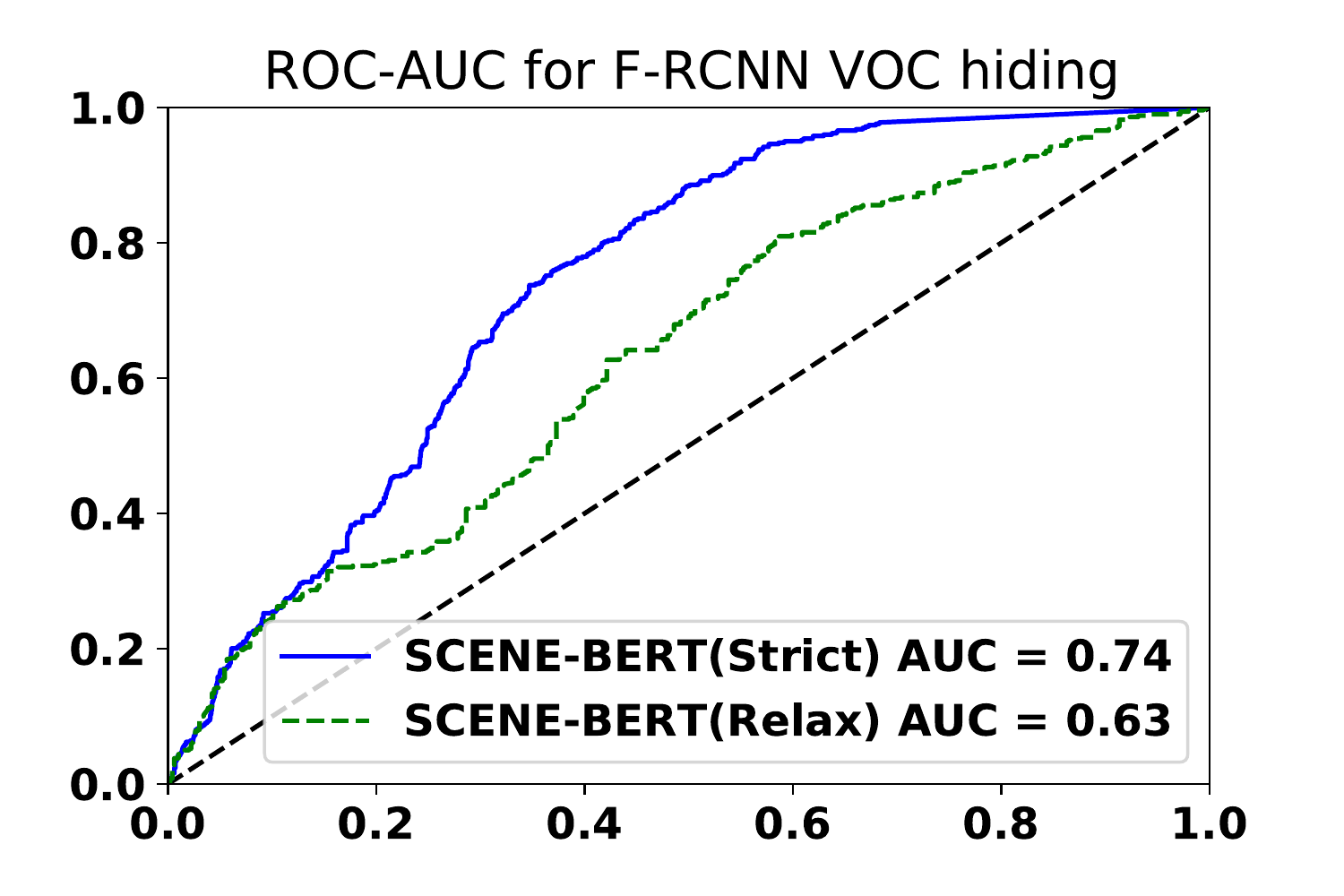}
    \end{subfigure}
    \begin{subfigure}{0.32\linewidth}
        \centering
        \includegraphics[width=\textwidth]{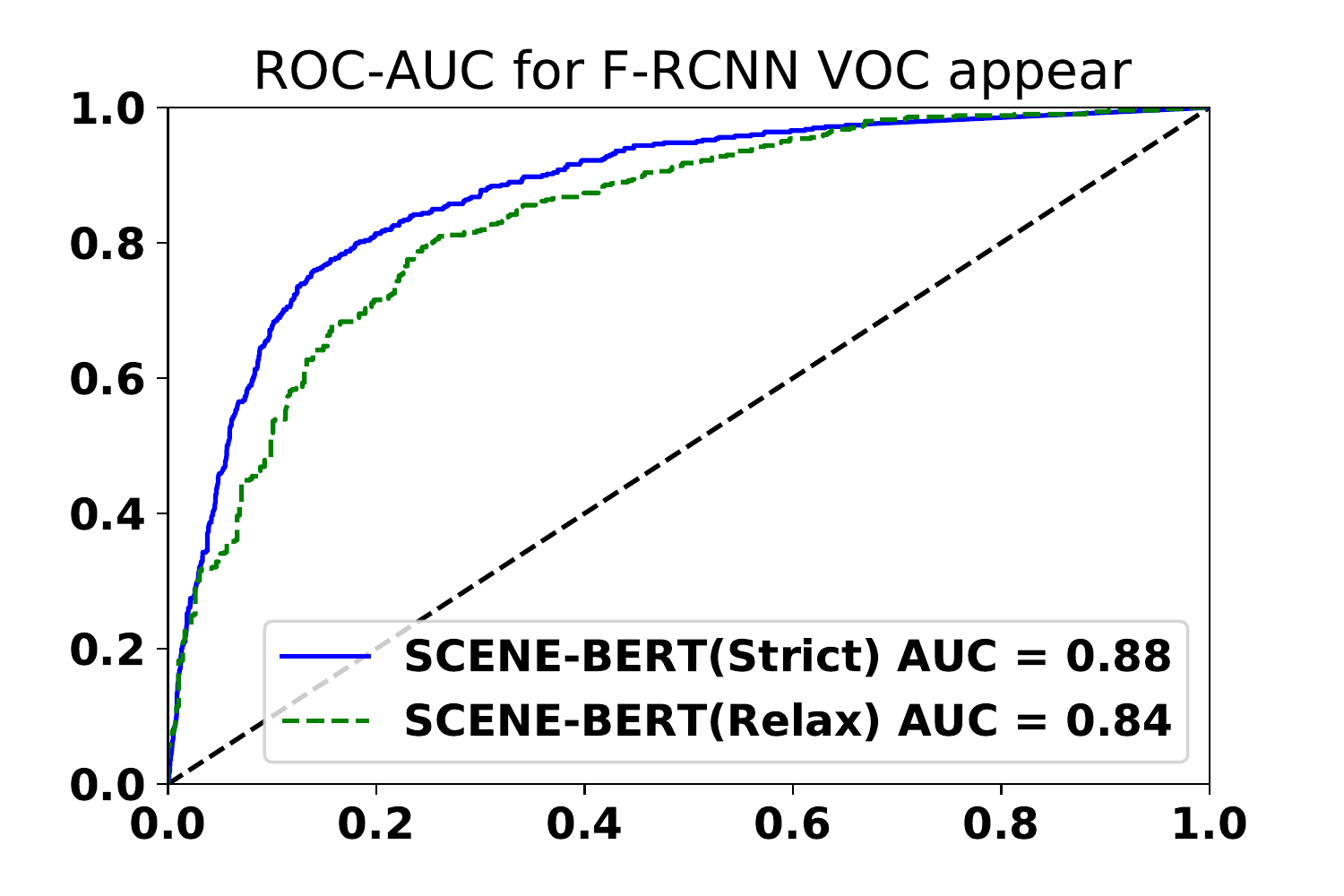}
    \end{subfigure}
    \caption{SCENE-BERT Strict vs. SCENE-BERT Relax on PASCAL VOC.
    } 
    \label{fig:strrel}
\end{figure*}

\PP{Effectiveness of Locations.}
To understand the importance of the coarse-grained location feature in our approach, we also performed the attack detection task with a relaxed consistency check,
where we only check the category and ignore the location when calculating the consistency score.
We name this approach SCENE-BERT Relax and the full version SCENE-BERT Strict.
The results are shown in~\autoref{fig:strrel}.
As we can see, the AUC is higher across all three types of attacks when we also check the coarse-grained location when calculating the consistency score.
We believe this shows (1) SCENE-BERT is able to capture location related dependencies between objects, and (2) even coarse-grained location information can help better detect the attacks. 

\subsection{Case Studies}
\begin{figure}[t]
    \centering
    \includegraphics[width=0.45\textwidth]{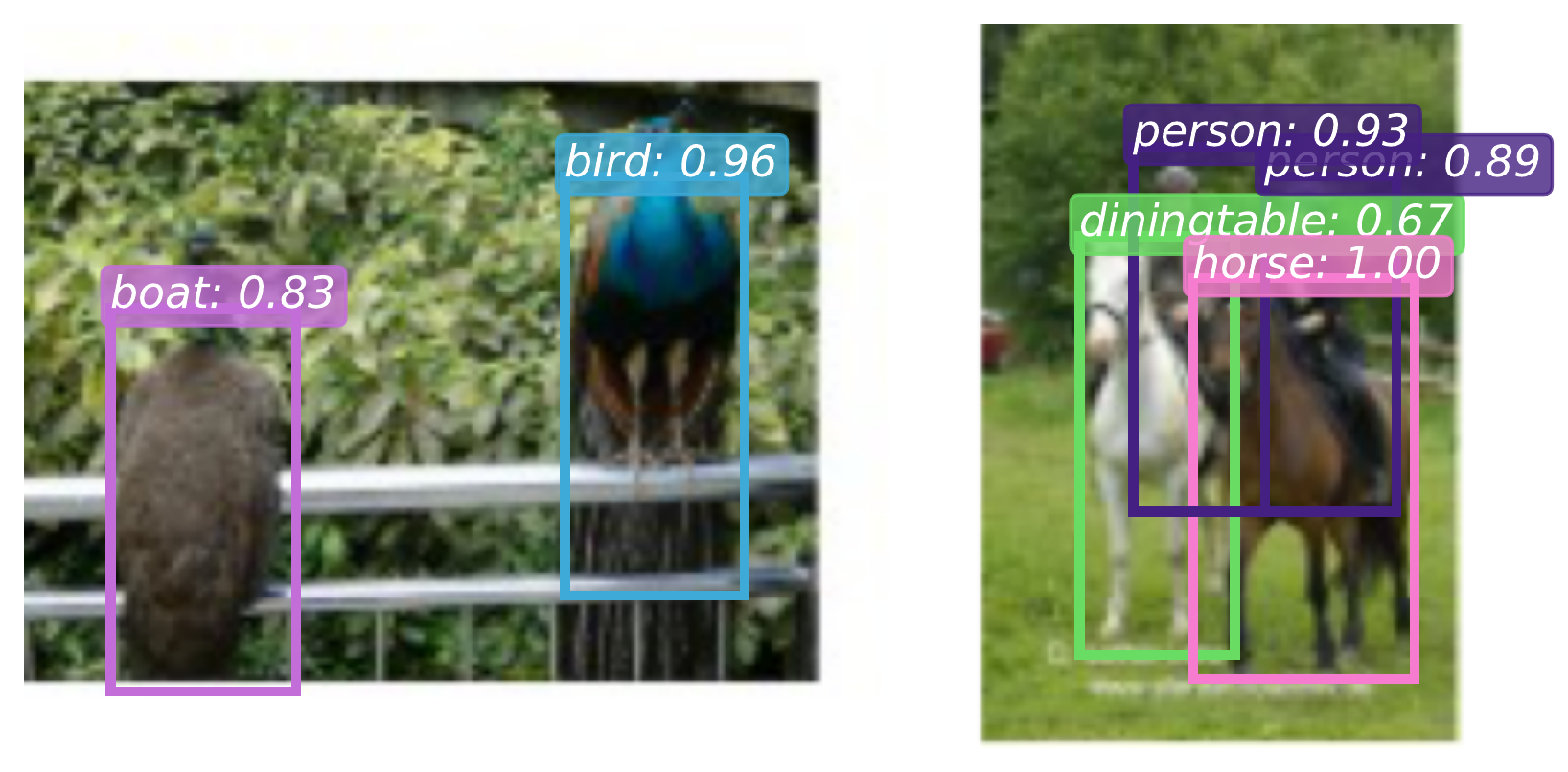}
    \caption{Examples where SCENE-BERT is able to detect the attack but SCEME does not.}
    \label{fig:showcase}
\end{figure}

While SCENE-BERT performed slightly worse than SCEME over the datasets, we also observed cases where SCENE-BERT can detect attacks that SCEME cannot.
\autoref{fig:showcase} shows two cases.
In the first case, the left bird (peacock) is perturbed to be a boat; and in the second case, the left horse is perturbed to be a dining table.
Both cases obviously violate the context consistency based on object co-occurrence hence were detected by SCENE-BERT.
We suspect that the reason that SCEME did not detect these attacks is because it also considers the visual features of the object and the auto-encoders may focus more on the visual features instead of the context.

{\color{black} While analyzing the evaluation results, we also noticed that if the attack is context consistent (e.g., misclassifies a bus to a car), then SCENE-BERT cannot detect such attacks. We want to argue that context consistency is one way to check for an attack and need not supplant, but can complement, other methods that can check individual objects or removal/addition of objects. Moreover, such context consistent attacks are likely to be less disruptive. For example, mis-clsassifying a bus to a car is unlikely to cause an autonomous vehicle to collide with the bus, but changing a speed limit sign in the middle of a road to a stop sign can lead to disastrous outcomes.
} 

%% file: tables/results.tex
\begin{table}[t]
    \centering
    \caption{Detection performance for F-RCNN, YOLO on VOC, COCO.\\
    (*This configuration is plotted in \autoref{fig:voc}, additional configurations reported in the supplementary material.)}
    \label{tab:result}
    \small
    \resizebox{8cm}{!}{
\begin{tabular}{lllccc}
\hline
\multirow{2}{*}{Dataset} & \multirow{2}{*}{Object Detector} & \multirow{2}{*}{Attack Detector} & \multicolumn{3}{c}{AUC}                                                              \\
                         &                                  &                                  & \multicolumn{1}{l}{Miscls} & \multicolumn{1}{l}{Hiding} & \multicolumn{1}{l}{Appear} \\ \hline
\multirow{5}{*}{VOC}     & \multirow{3}{*}{F-RCNN*}          & SCENE-BERT                       & 0.88                       & 0.74                       & 0.88                       \\
                         &                                  & SCEME                            & 0.93                       & 0.95                       & 0.87                       \\
                         &                                  & Feature Squeeze                  & 0.53                       & 0.52                       & 0.52                       \\ \cline{2-6} 
                         & \multirow{2}{*}{YOLO}            & SCENE-BERT                       & 0.89                       & 0.74                       & 0.90                       \\
                         &                                  & Feature Squeeze                  & 0.77                       & 0.75                       & 0.79                       \\ \hline
\multirow{4}{*}{COCO}    & \multirow{2}{*}{F-RCNN}          & SCENE-BERT                       & 0.84                       & 0.55                       & 0.85                       \\
                         &                                  & Feature Squeeze                  & 0.60                       & 0.74                       & 0.60                       \\ \cline{2-6} 
                         & \multirow{2}{*}{YOLO}            & SCENE-BERT                       & 0.86                       & 0.55                       & 0.88                       \\
                         &                                  & Feature Squeeze                  & 0.66                       & 0.60                       & 0.67                       \\ \hline
\end{tabular}
}

\end{table}

%% file: 5conclu.tex
\section{Conclusion}

Motivated by the observation that language descriptions of a natural scene images have captured the object co-occurrence relationship,
we propose using a language model to learn the dependencies between objects and using the trained model to perform context consistency checks to detect adversarial attacks.
Compared to previous context-consistency-based detection method, our approach can be paired with most object detectors and does not require modification or retraining to the object detector.
Our experiments show that our method is very effective in detecting a variety of attacks on two large scale datasets:
it significantly outperforms a state-of-the-art context-agnostic method and is comparable to previous context-aware method that is model-dependent.

%% file: 6supp.tex

In the supplementary material, we conduct case study on attacks that can bypass our context consistency checker and benign images that are detected as adversarial (Section~\ref{sec:supp_case_study}).
We present the implementation details about the baseline models (Feature Squeeze and SCEME) used in the paper.

\begin{figure}[ht]
    \includegraphics[width=0.5\textwidth]{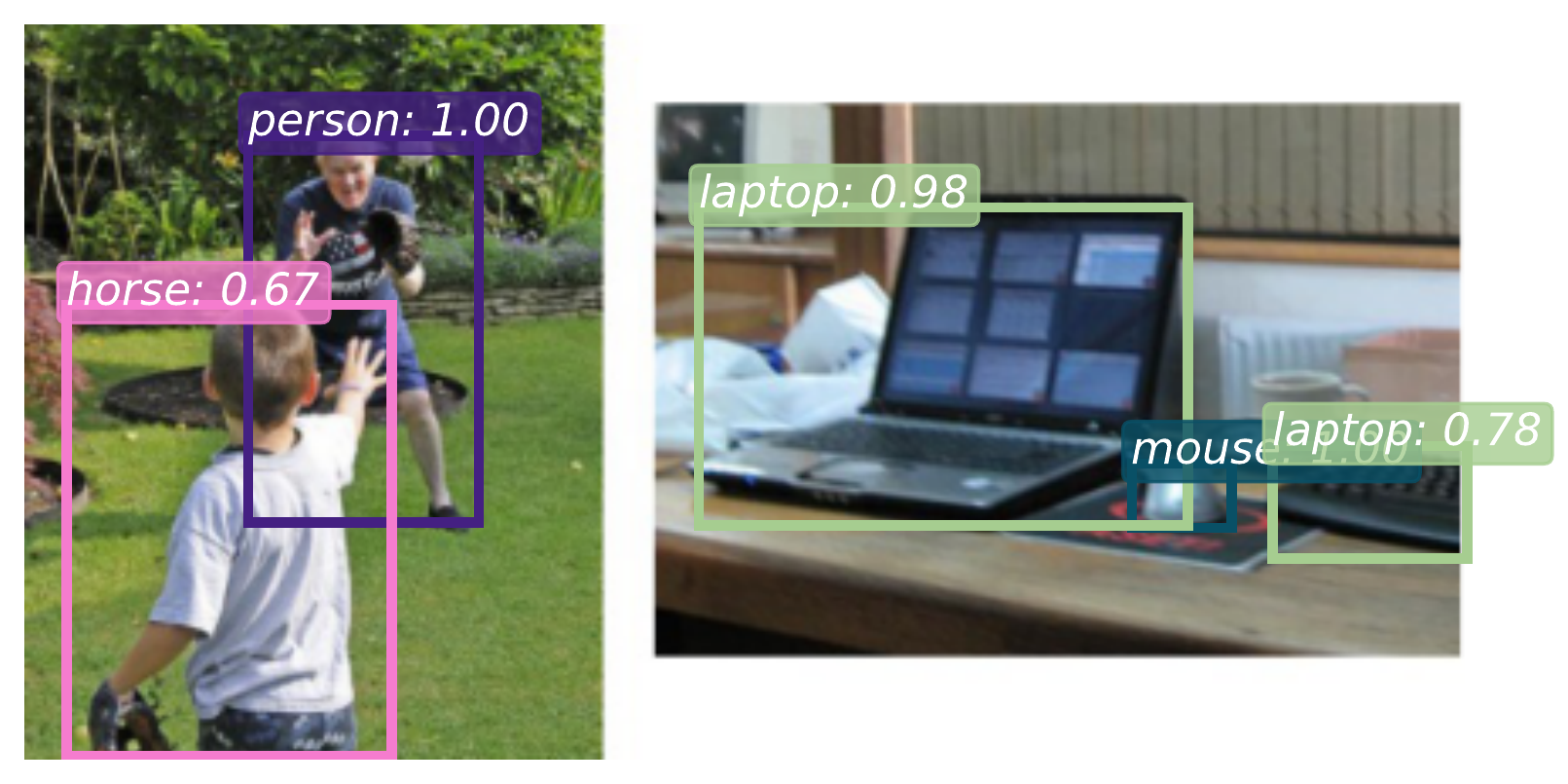}
    \caption{The adversarial examples that our proposed SCENE-BERT model cannot detect.}
    \label{fig:supp_showcase}
\end{figure}

\begin{figure}[ht]
    \includegraphics[width=0.5\textwidth]{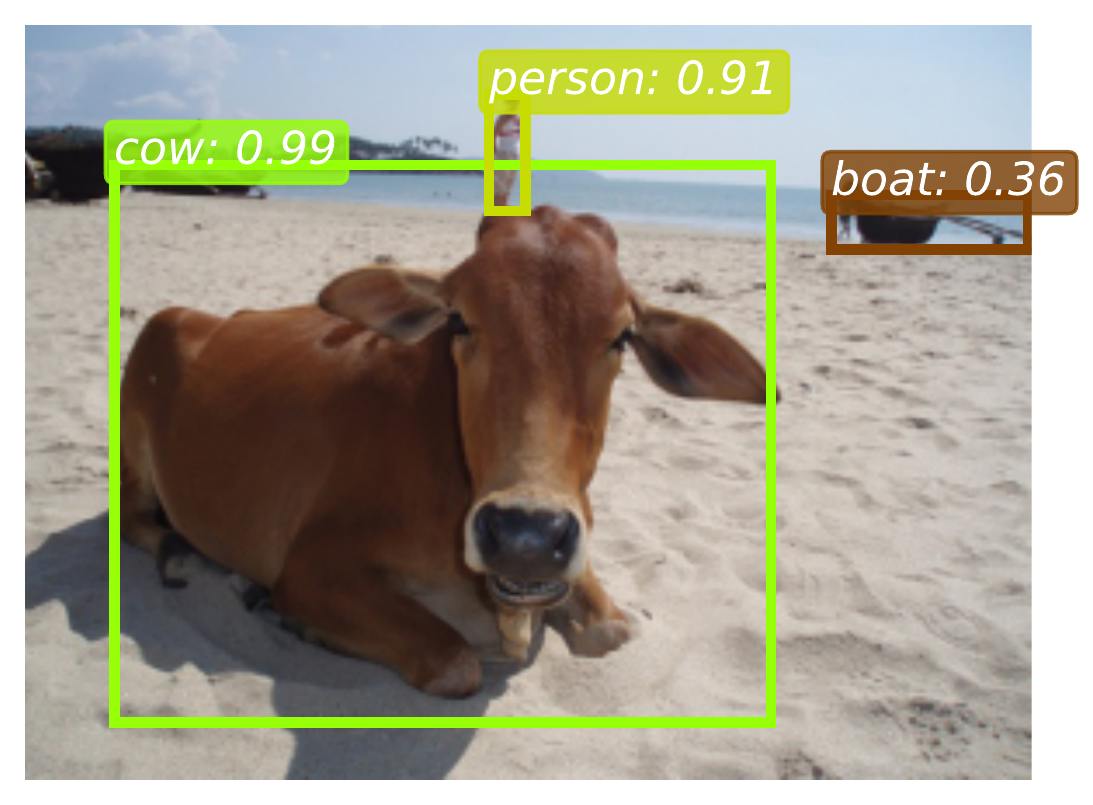}
    \caption{The benign example that our proposed SCENE-BERT model detect as adversarial.}
    \label{fig:supp_showcase_fp}
\end{figure}

\begin{figure}[ht]
    \includegraphics[width=0.5\textwidth]{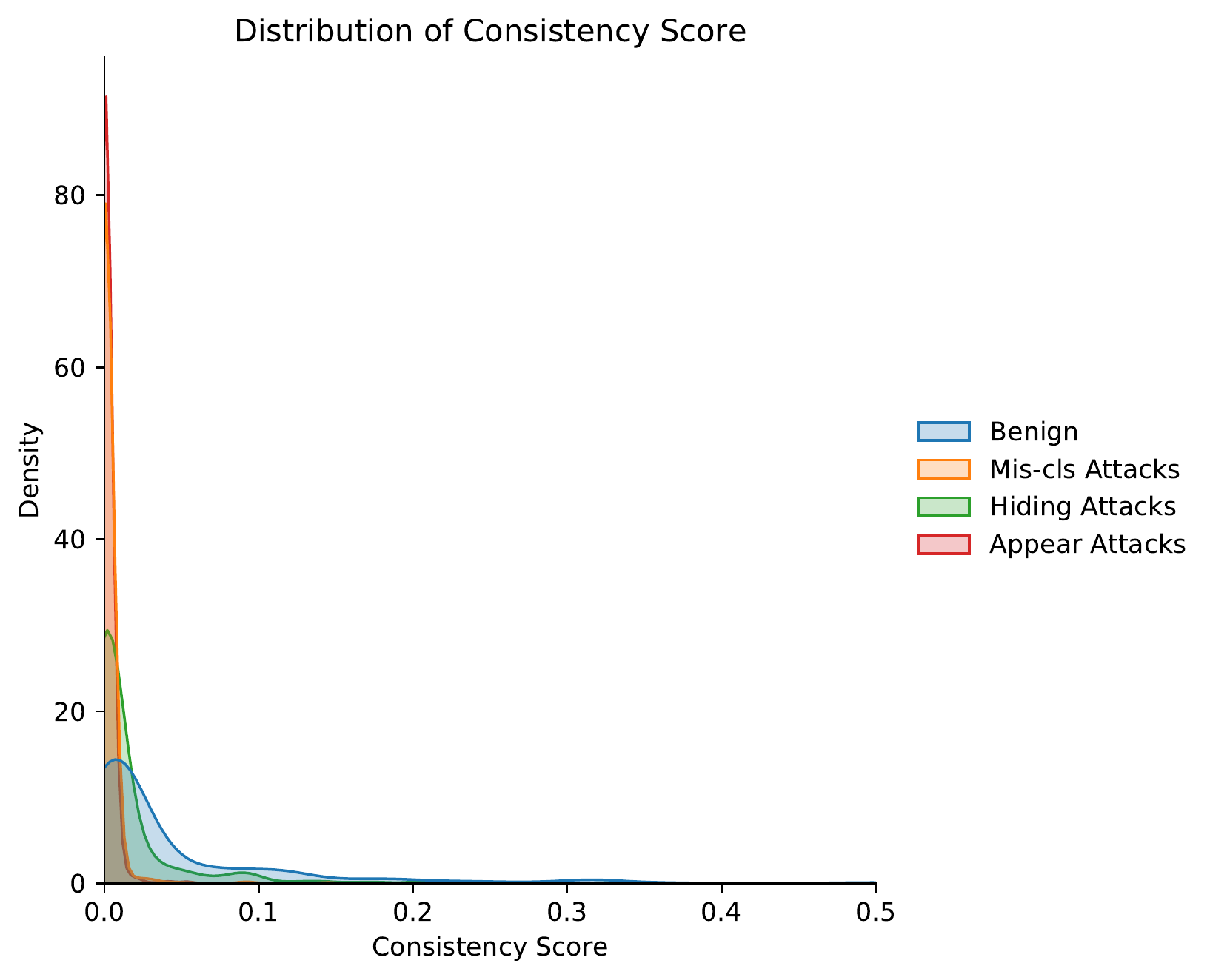}
    \caption{The distribution of consistency score for three types of attacks.}
    \label{fig:supp_dist}
\end{figure}

\section{Case Study}
\label{sec:supp_case_study}

In this section, we first show (1) attack examples that cannot be detect by our language-based consistency checker SCENE-BERT and (2) benign examples that are reported as adversarial.
Then we present the distribution of consistency scores.

\textbf{False Negatives.}
For the adversarial image on left hand side of~\autoref{fig:supp_showcase}, the attack goal is to misclassify the person into a horse.
Because the (misclassified) detection result will be described ``a person (and) a horse,'' which is common, our context-consistency-based detector cannot tell the image has been perturbed.
For the adversarial image on right hand side of~\autoref{fig:supp_showcase}, the attack goal is to misclassify the rightmost keyboard into a laptop.
Similarly, because it is ordinary for a laptop to co-occur with another laptop, our approach cannot detect the image as perturbed.

\textbf{False Positives.}
There are also benign examples that are detected as adversarial by SCENE-BERT. \autoref{fig:supp_showcase_fp} shows an example. In this case, a person lying on top of a cow is common, but a boat co-occur with cow is rare in the training set or has never been seen before. In this case, though the object boat is indeed an out-of-context anomaly, it is not an attack.

\textbf{Density Graph.}
To understand how common such cases are, we plotted the distribution of consistency score for the three types of attacks and the benign cases in~\autoref{fig:supp_dist}.
As we can see, the consistency score for misclassification appear attacks mainly concentrate at around 0, while the majority of the benign images have higher consistency score.
Hiding attacks are hard to detect because hiding attacks usually do not violate context-consistency (e.g., hide one person from a group of person will not cause the context change too much).
However, hiding attacks are still detectable in subtle cases.
For instance, a watch should be wear by a person or located on a table, if the person are hidden and not table detected, then the consistency can be considered attacked.

These false negatives and false positives are caused by a fundamental limitation of our consistency-based attack detection approach---if the attack itself is context-aware, then we cannot use context-consistency to detect such attacks; on the other hand, if a benign case has never been seen before, then we may also report it as attacks.
However, we argue that (1) false positives can be reduced by extending the training set (e.g., by using natural language datasets),
(2) as context imposes additional constraints, constructing context-aware attacks are likely to be more expensive; and
(3) more importantly, context-consistent attacks may cause be able to lead to dire consequences (e.g., misclassifying STOP sign to YIELD sign may not lead to traffic accidents).

\section{Implementation Details}

\subsection {Feature Squeeze}\label{sec:supp_fs}
In this subsection, we explain how the baseline Feature Squeeze is implemented.
Algorithm~\ref{algo:fs} shows the algorithm. $\mathbf{Img_{O}}$ denotes the original image, $\mathbf{Img_{Q}}$ defined in line 1 denotes the quantized image after squeezing. $\mathbf{PR_{O}}$ defined in line 2 denotes the prediction result for the original image from object detector $g(\cdot)$, while $\mathbf{PR_{FS}}$ defined in line 3 denotes the prediction result of quantized image. $R_{O}$ and $R_{FS}$ defined in line 7, 9 denote one region from the prediction result for original image and quantized image respectively. Note that we take the highest distance among all regions as a represent to the distance of whole image. Furthermore, for each region, we take the lowest distance calculating from all its overlapped as the distance for the queried region. Under the consideration that we usually take the category with highest confidence score for the regions dumped from object detector, we manually set rule, i.e. the distance is $1$ for regions with different predicted categories.

\begin{algorithm}[ht]
	\fontsize{8}{8}
	\selectfont
	\SetAlgoLined
	\SetKwInOut{Input}{Input}\SetKwInOut{Output}{Output}
	
	\Input{Image to be tested $\mathbf{Img_{O}}$,\\ the quantilize function $f(\cdot)$,\\ the object detector $g(\cdot)$}
	\Output{Distance $d$}
    $\mathbf{Img_{Q}} = f(\mathbf{Img_{O}})$ \\
    $\mathbf{PR_{O}} = g(\mathbf{Img_{O}})$ \\
    $\mathbf{PR_{FS}} = g(\mathbf{Img_{Q}})$ \\
	$d = 0$ \\
    $n = \text{length of PR{FS}}$ \\
	\For {$i=1 \text{ to } n $}{
        $R_{FS} = \mathbf{PR_{FS}[i]}$ \\
        $minDistance = 1$ \\
        \For { $R_{O} \text{ in } getOverlap( R_{FS}, \mathbf{PR_{O}})$ }{
            $distance = getDistance(R_{FS}, R_{O})$ \\
            $minDistance = \min(distance, minDistance)$ \\
        }
        $d = \max(minDistance, d)$ \\
	}
    \If {$R_{O}\text{ in }\mathbf{PR_{O}} \text{ overlap with nothing in } \mathbf{PR_{FS}}$} {
        $d = 1$
    }
	\Return $d$
	\caption{Calculate the distance using Feature Squeeze of an image.}
	\label{algo:fs}
\end{algorithm}

\subsection {SCEME} \label{sec:supp_sceme}
In this subsection, we explain how the baseline SCEME model is implemented.
Algorithm~\ref{algo:sceme} illustrates how we adapted the original SCEME, which works at region proposal level, to make it work at whole image level. $\mathbf{PR}$ defined in line 2 denotes the prediction results. $CP$ defined in line 6 denotes the Context Profile extracted from intermediate layer of F-RCNN which will be passed to SCEME model to generate a reconstruction error. We take the highest reconstruction error as the reconstruction error of the whole image.

\begin{algorithm}[ht]
	\fontsize{8}{8}
	\selectfont
	\SetAlgoLined
	\SetKwInOut{Input}{Input}\SetKwInOut{Output}{Output}
	
	\Input{Image to be tested $\mathbf{Img}$, \\
    the F-RCNN object detector $f(\cdot)$, \\
    the trained SCEME function $g(\cdot,\cdot)$,
    }
	\Output{Reconstruction Error $e$}
	$e = 0$ \\ 
    $\mathbf{PR} = f(\mathbf{Img})$ \\
    $n = $ length of $\mathbf{PR}$ \\
	\For {$i=1 \text{ to } n $}{
		$Category, CP = \mathbf{PR}[i]$ \\
		$rec = g(Category, CP)$ \\
		$e = \max(e, rec)$ \\
	}
	\Return $e$
	\caption{Calculate the SCEME reconstruction error at image level.}
	\label{algo:sceme}
\end{algorithm}


%
Note that because the categories in MS COCO dataset are biased and SCEME requires on auto-encoder for each category, it cannot be trained well on categories that rarely occur.
For this reason, we did not measure SCEME's performance on the MS COCO dataset.

\begin{figure*}[t]
    \centering
    \begin{subfigure}{0.32\linewidth}
        \centering
        \includegraphics[width=\textwidth]{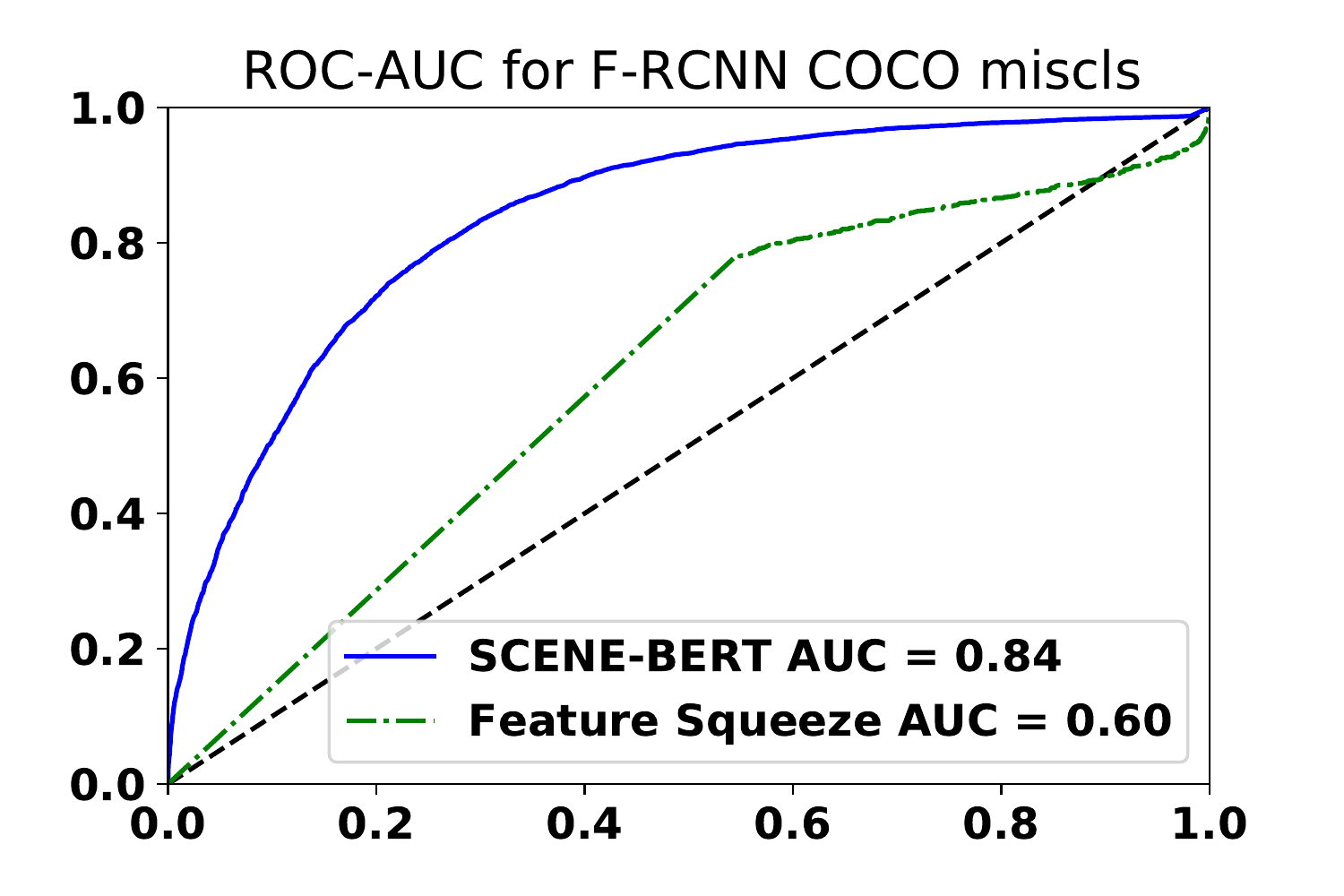}
    \end{subfigure}
    \begin{subfigure}{0.32\linewidth}
        \centering
        \includegraphics[width=\textwidth]{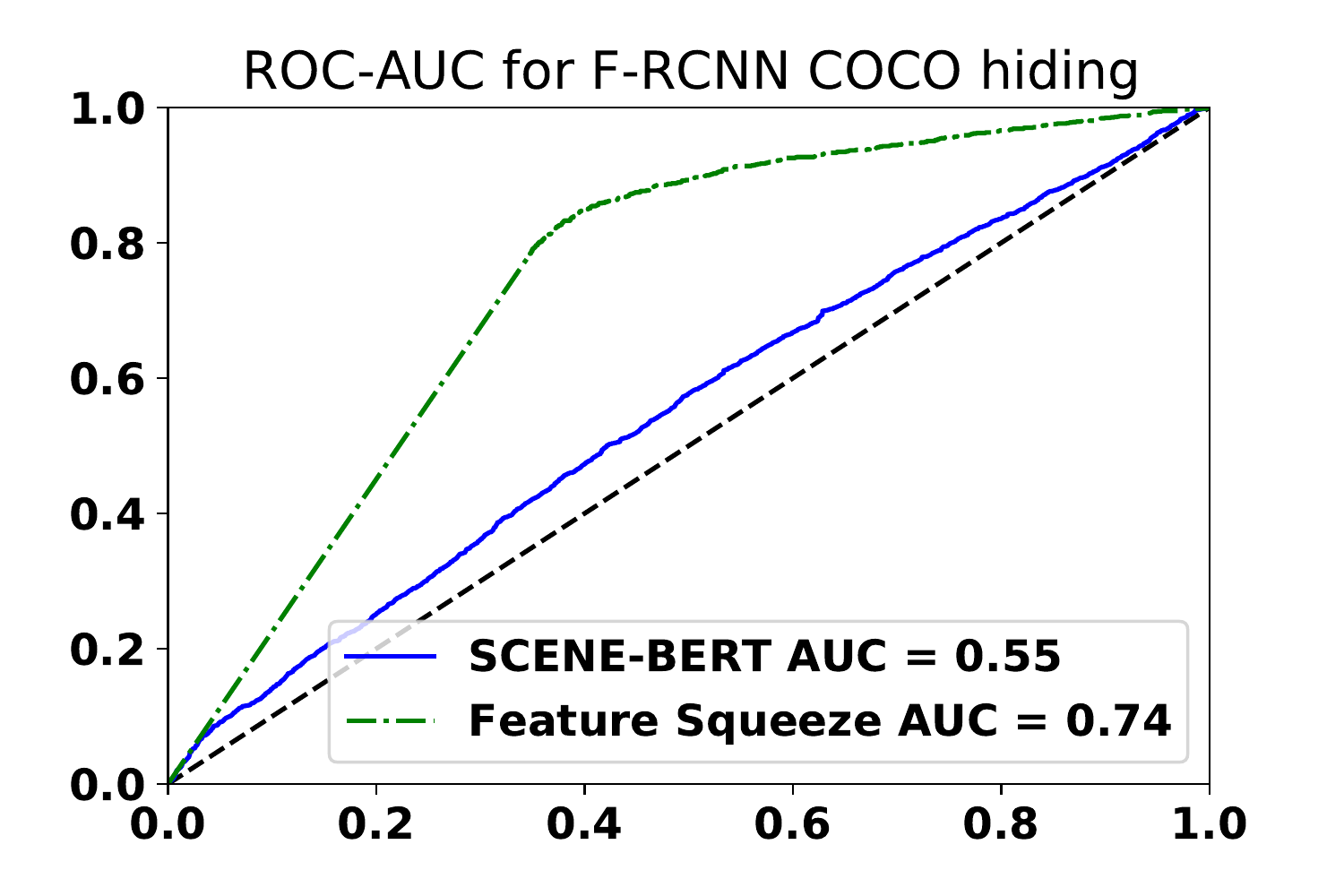}
    \end{subfigure}
    \begin{subfigure}{0.32\linewidth}
        \centering
        \includegraphics[width=\textwidth]{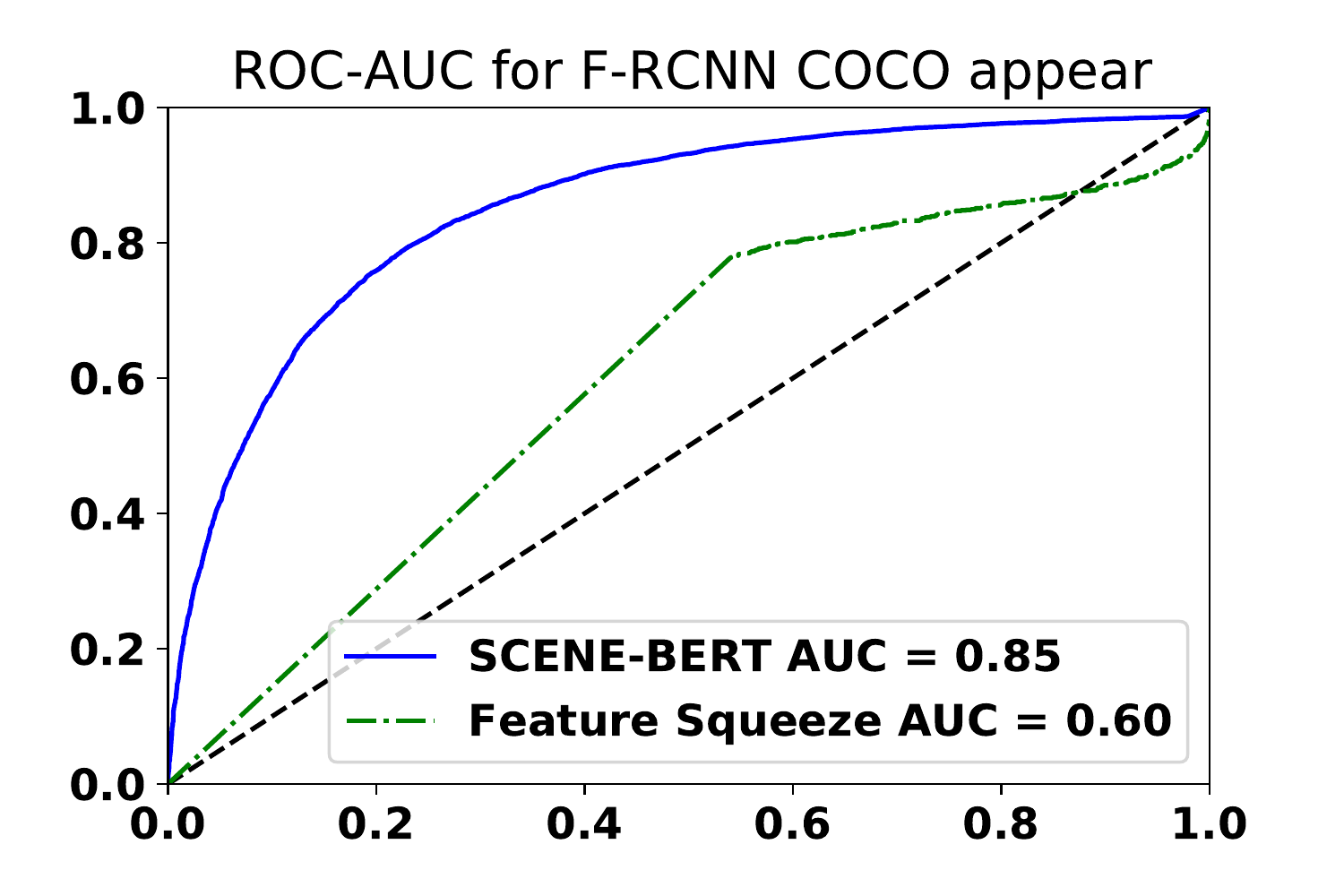}
    \end{subfigure}
    \caption{ROC-AUC for F-RCNN on COCO} 
    \label{fig:frcnncoco}
\end{figure*}
\begin{figure*}[t]
    \centering
    \begin{subfigure}{0.32\linewidth}
        \centering
        \includegraphics[width=\textwidth]{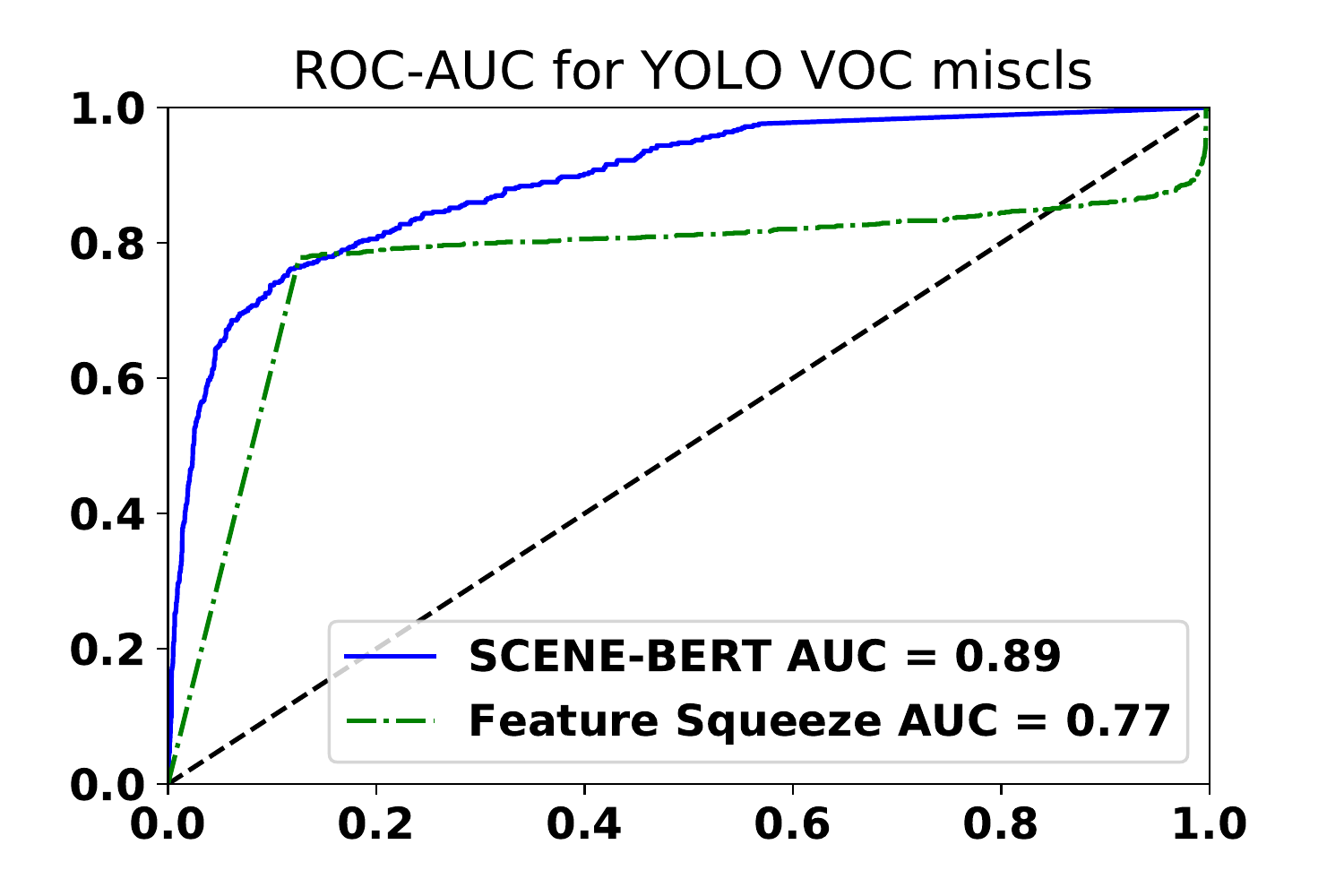}
    \end{subfigure}
    \begin{subfigure}{0.32\linewidth}
        \centering
        \includegraphics[width=\textwidth]{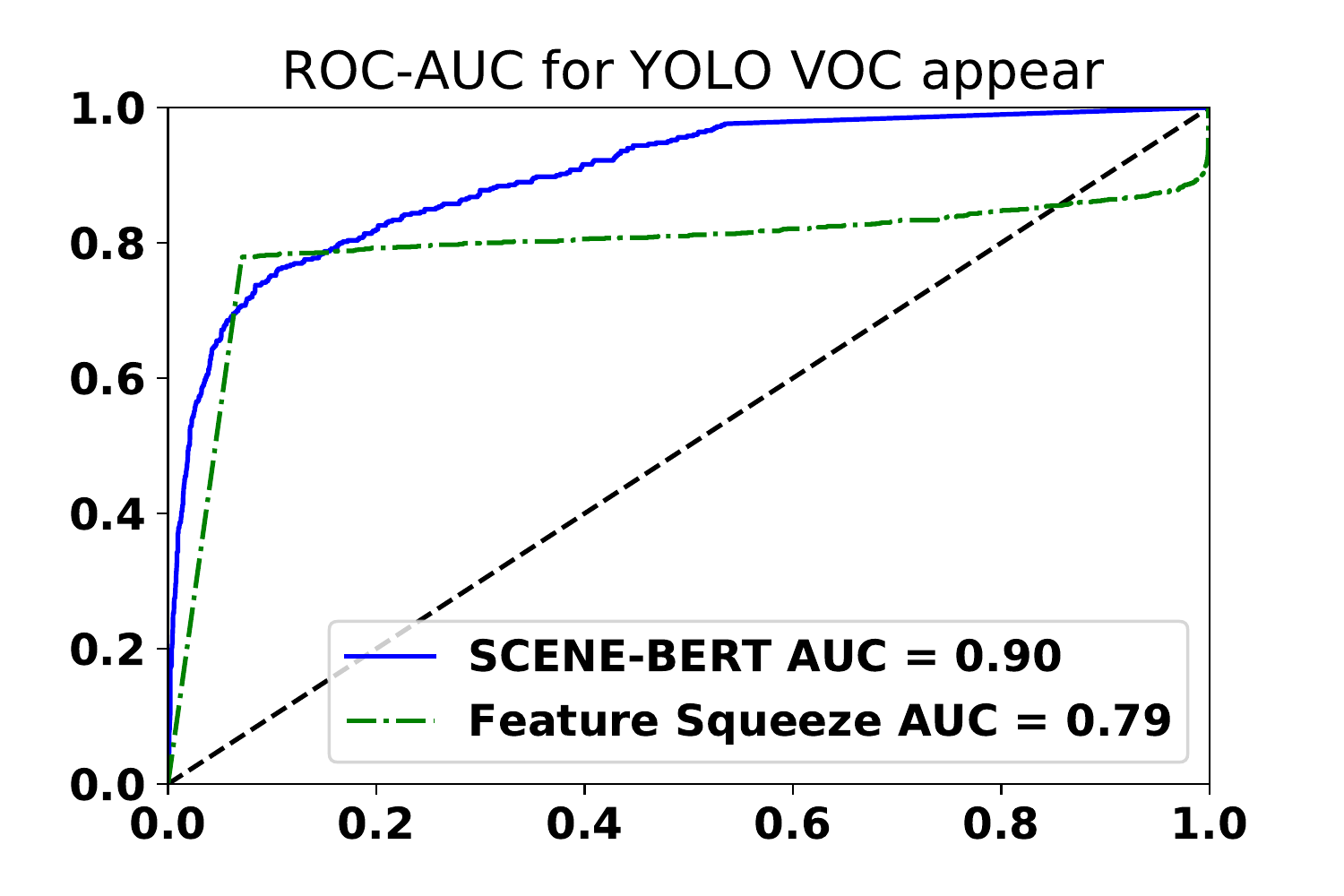}
    \end{subfigure}
    \begin{subfigure}{0.32\linewidth}
        \centering
        \includegraphics[width=\textwidth]{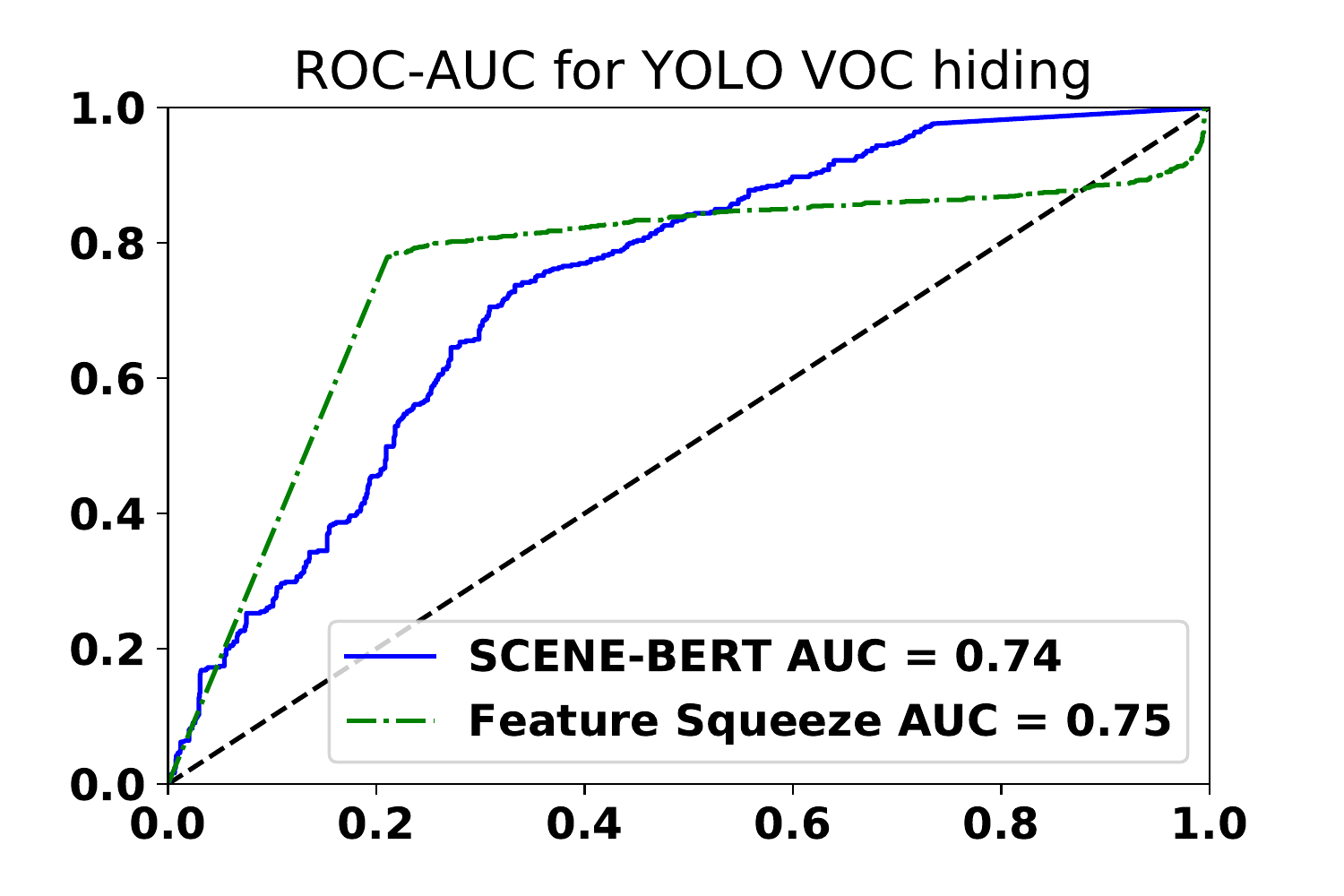}
    \end{subfigure}
    \caption{ROC-AUC for YOLO on COCO} 
    \label{fig:yolococo}
\end{figure*}
\begin{figure*}[t]
    \centering
    \begin{subfigure}{0.32\linewidth}
        \centering
        \includegraphics[width=\textwidth]{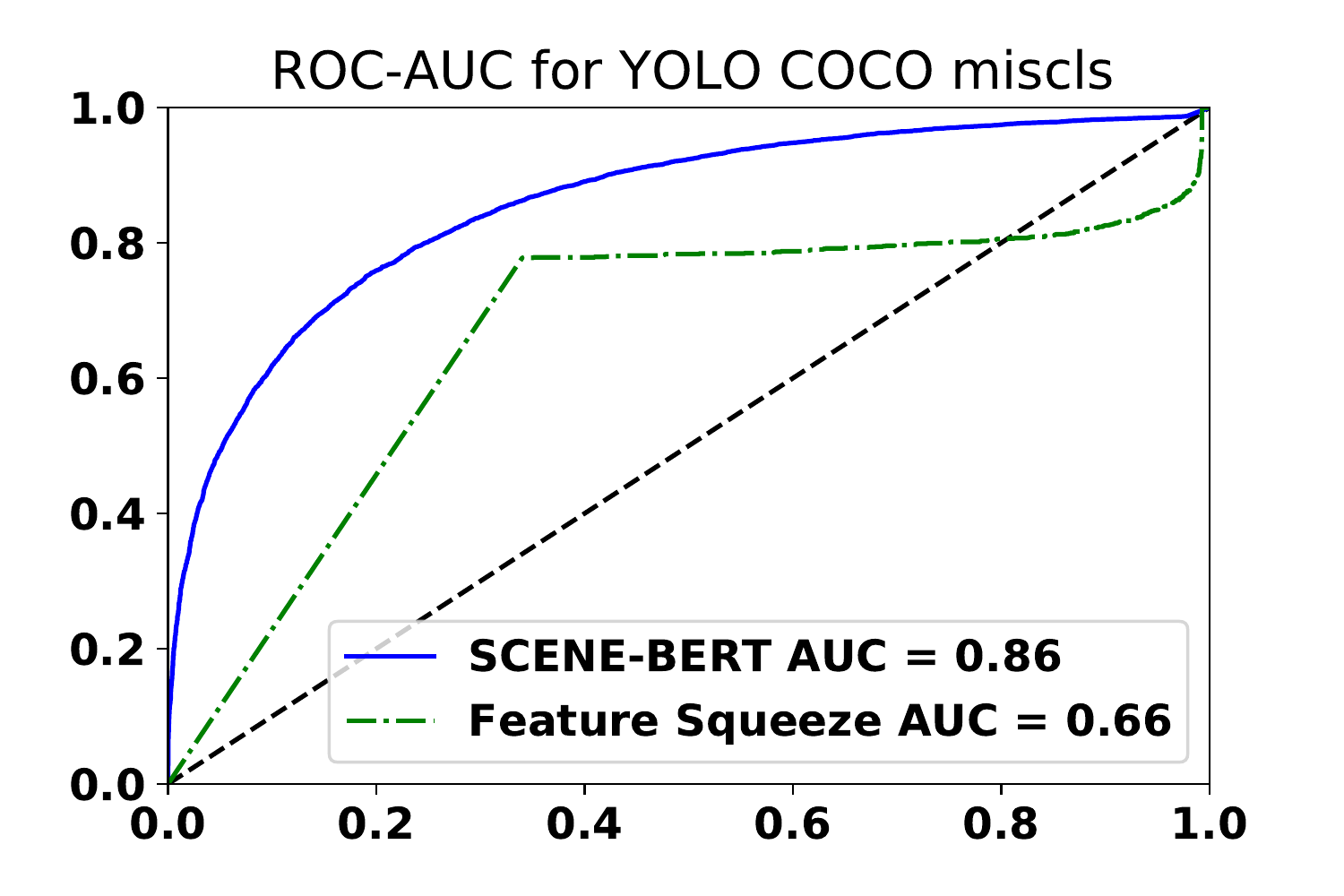}
    \end{subfigure}
    \begin{subfigure}{0.32\linewidth}
        \centering
        \includegraphics[width=\textwidth]{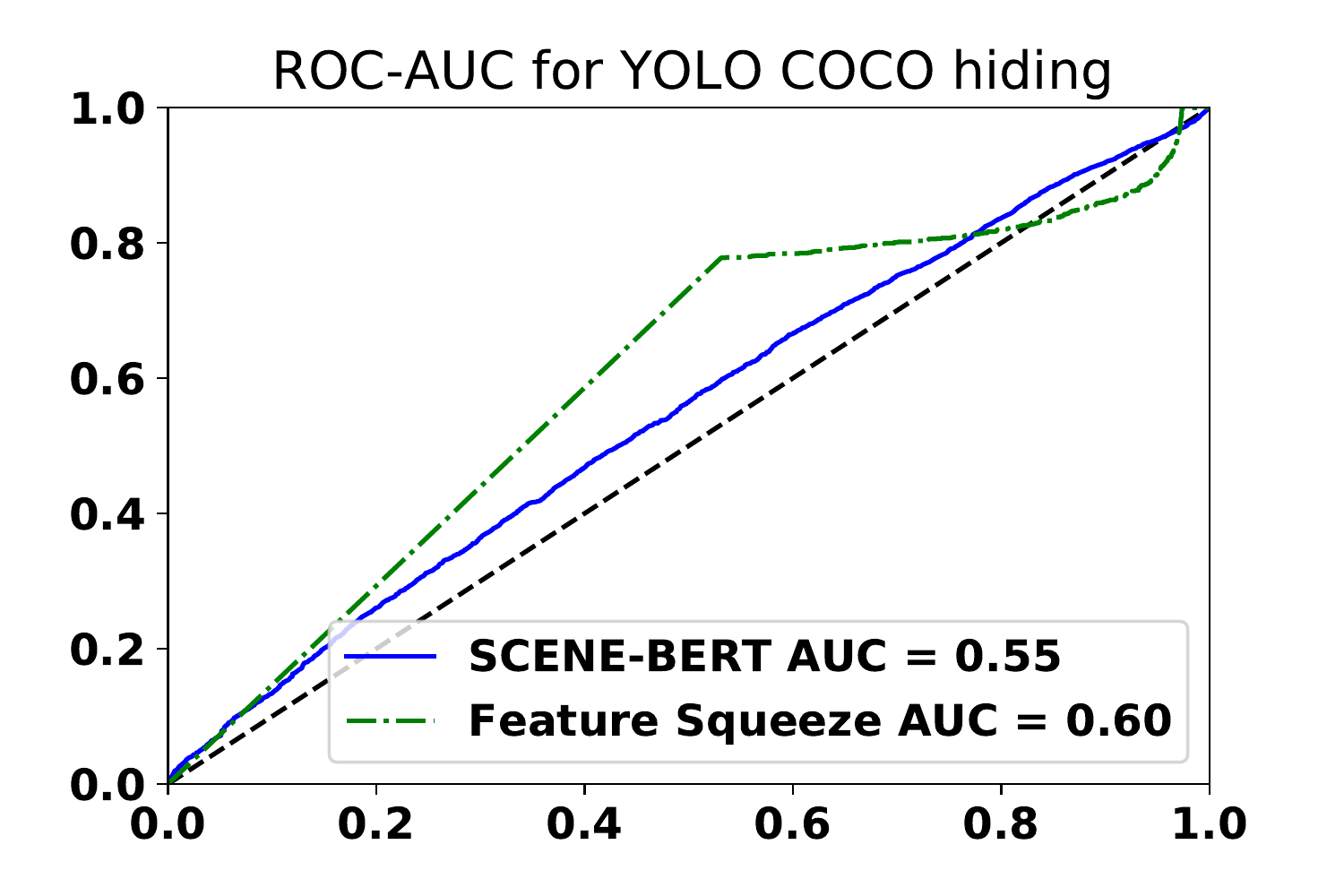}
    \end{subfigure}
    \begin{subfigure}{0.32\linewidth}
        \centering
        \includegraphics[width=\textwidth]{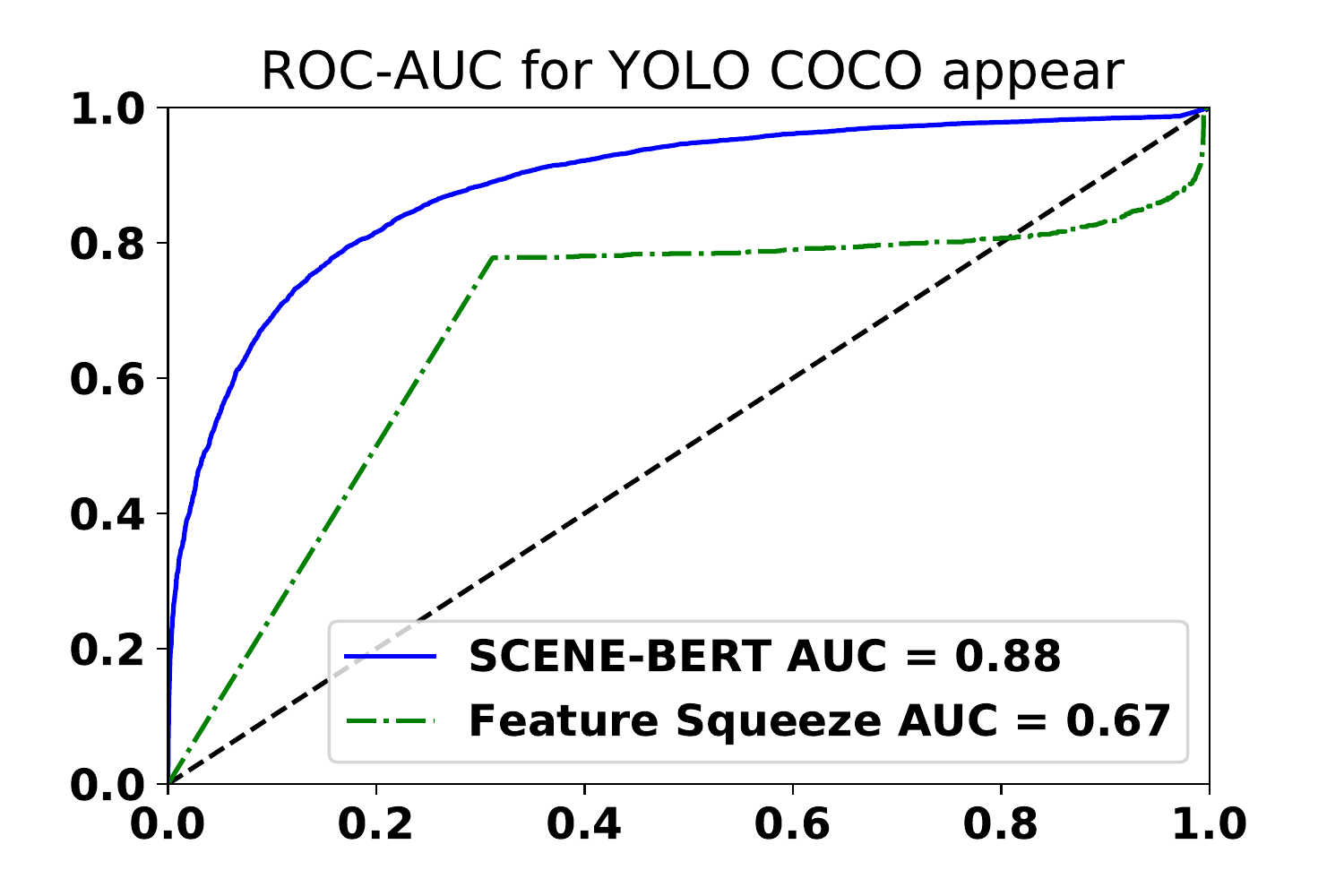}
    \end{subfigure}
    \caption{ROC-AUC for YOLO on VOC} 
    \label{fig:yolovoc}
\end{figure*}
\section{Additional Results}

We plot the ROC-AUC curve for F-RCNN on COCO, YOLO on VOC, and YOLO on COCO in \autoref{fig:frcnncoco}, \autoref{fig:yolovoc}, and \autoref{fig:yolococo}.
\clearpage